# Kernel Methods on Approximate Infinite-Dimensional Covariance Operators for Image Classification

Hà Quang Minh*, Marco San Biagio*, Loris Bazzani, and Vittorio Murino

**Abstract**—This paper presents a novel framework for visual object recognition using infinite-dimensional covariance operators of input features in the paradigm of kernel methods on infinite-dimensional Riemannian manifolds. Our formulation provides in particular a rich representation of image features by exploiting their non-linear correlations. Theoretically, we provide a finite-dimensional approximation of the Log-Hilbert-Schmidt (Log-HS) distance between covariance operators that is scalable to large datasets, while maintaining an effective discriminating capability. This allows us to efficiently approximate any continuous shift-invariant kernel defined using the Log-HS distance. At the same time, we prove that the Log-HS inner product between covariance operators is only approximable by its finite-dimensional counterpart in a very limited scenario. Consequently, kernels defined using the Log-HS inner product, such as polynomial kernels, are not scalable in the same way as shift-invariant kernels. Computationally, we apply the approximate Log-HS distance formulation to covariance operators of both handcrafted and convolutional features, exploiting both the expressiveness of these features and the power of the covariance representation. Empirically, we tested our framework on the task of image classification on twelve challenging datasets. In almost all cases, the results obtained outperform other state of the art methods, demonstrating the competitiveness and potential of our framework.

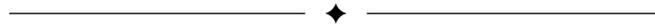

## 1 INTRODUCTION

Covariance descriptors are a powerful image representation approach in computer vision. In this approach, an image is compactly represented by a covariance matrix encoding correlations between different features extracted from that image. This representation has been demonstrated to work very well in numerous vision tasks, including tracking [42], object detection and classification [50], [51], and image retrieval [11]. Covariance descriptors, properly regularized if necessary, are symmetric positive definite (SPD) matrices, which do not form a vector subspace of Euclidean space under the standard matrix addition and scalar multiplication operations, but form a Riemannian manifold. The optimal measure of similarity between covariance descriptors is thus not the Euclidean distance, but a metric that captures this manifold structure. One of the most commonly used Riemannian metrics in the literature is the Log-Euclidean metric developed by [2]. This is a so-called *bi-invariant Riemannian metric* under which the manifold is flat, that is having zero curvature. Therefore,

- *These two authors contributed equally to this work.
- H.Q. Minh is with the Department of Pattern Analysis and Computer Vision (PAVIS), Istituto Italiano di Tecnologia (IIT), Genova 16163, ITALY. E-mail: minh.haquang@iit.it
- M. San Biagio was with the Department of Pattern Analysis and Computer Vision (PAVIS), Istituto Italiano di Tecnologia (IIT), Genova 16163, ITALY, at the time this work was carried out. E-mail: marco.sanbiagio@gmail.com
- L. Bazzani was with the Department of Computer Science, Dartmouth College, Hanover (NH) 03755, USA, at the time this work was carried out. E-mail: loris.bazzani@gmail.com
- V. Murino is with the Department of Pattern Analysis and Computer Vision (PAVIS), Istituto Italiano di Tecnologia (IIT), Genova 16163, and the Department of Computer Science, University of Verona, ITALY. E-mail: vittorio.murino@iit.it

it is efficient to compute and can be used to define many positive definite kernels, allowing kernel methods to be applied directly on the manifold. This latter property has been exploited successfully in various recent work in computer vision [22], [30]. However, a major limitation of covariance matrices is that they only capture *linear* correlations between input features.

In this work, we propose to use *infinite-dimensional covariance operators* as image representations. These are covariance matrices of infinite-dimensional features, which are induced implicitly when a positive definite kernel ($K_1$ in Fig. 1), such as the Gaussian kernel, is applied to the original image features. These covariance operators capture in particular *non-linear correlations* between the original input features. Each image is then represented by one such covariance operator.

For tasks such as image classification, we require the notion of similarity between image representations, which in this case means the similarity between the corresponding covariance operators. It is known that covariance operators, properly regularized, lie on the *infinite-dimensional Riemannian manifold* of positive definite operators [28]. On this manifold, the generalization of the Log-Euclidean metric is the *Log-Hilbert-Schmidt* (Log-HS) metric [36]. Give two covariance operators, as a similarity measure, we can compute either the Log-HS distance or the Log-HS inner product between them. Having computed the Log-HS distances or Log-HS inner products between the covariance operators, another positive definite kernel ($K_2$ in Fig. 1) can be computed using these distances or inner products and used as input to a kernel classifier, e.g. SVM.

In essence, the above two steps, namely image representation by covariance operators and kernel classification, together make up a two-layer kernel machine (Fig. 1). The kernel $K_1$ in the first layer, with the low-level image features as input, induces covariance operators which capture non-linear correlations between these features. The kernel $K_2$ in the second layer, with the Log-HS distances or Log-HS inner products between covariance operators as input, allows the application of kernel methods, e.g. SVM, to these operators. The result of this double kernelization process is a more powerful representation that can better capture the expressiveness of the image features by exploiting both the power of kernel methods and the Riemannian manifold setting of covariance operators.

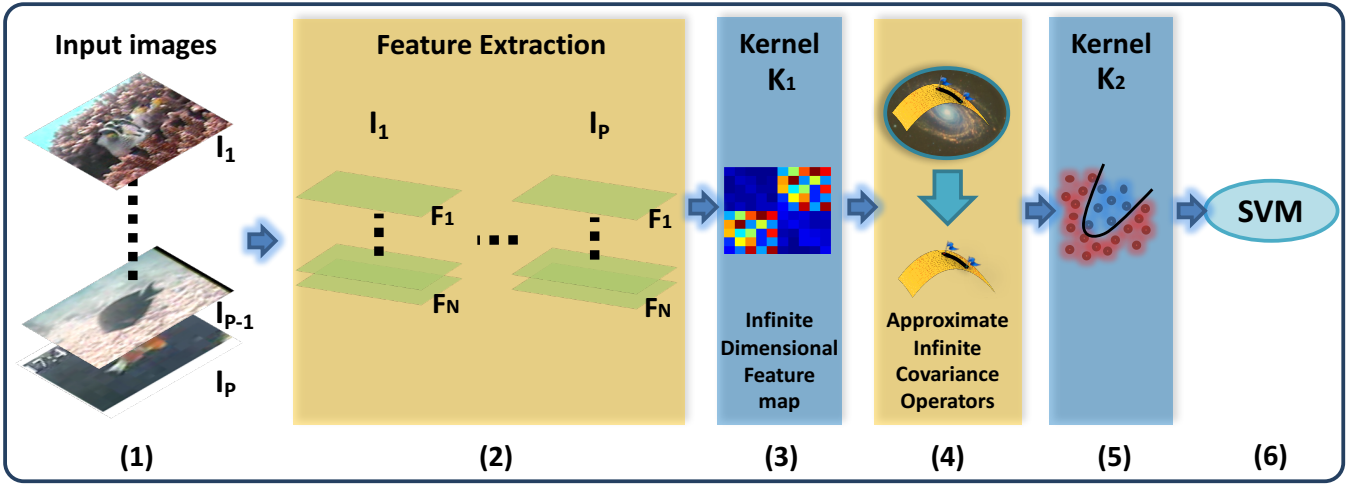

Fig. 1. Model of the proposed framework. Given a set of images $\{I_1, \ldots, I_P\}$ (step 1), $N$ (low-level) features $\{F_1, \ldots F_N\}$ are extracted from each image (step 2). Next, a positive definite kernel ($K_1$) is defined (step 3), which *implicitly* maps the features $F_i$'s into an infinite-dimensional feature space, *implicitly* defining an infinite-dimensional covariance operator corresponding to each image. We then compute, *explicitly*, a finite-dimensional approximation for each of these covariance operators (step 4). Finally, another positive definite kernel ($K_2$) is computed using the distances between the approximate covariance operators (step 5) and used as input to an SVM classifier (or any other kernel method) (step 6).

However, as with many kernel methods, one drawback of the original Log-HS metric formulation in [36] is that it tends not to scale well to large datasets.

**Approximate Log-HS distance**. To carry out the above kernelization efficiently, we develop the following novel mathematical and computational framework. In this paper, we propose an *approximate Log-HS* distance formulation. This is done by approximating the implicit infinite-dimensional covariance operators above by explicit finite-dimensional covariance matrices, which are computed using explicit approximate feature maps of the original kernel ($K_1$ in Fig. 1). We demonstrate that the approximate Log-HS distance is substantially faster to compute than the true Log-HS distance, with relatively little loss in the performance of the resulting algorithm. *The first main theoretical contribution of the present work is the mathematical derivation and justification of the approximate Log-HS distance.* In particular, we provide the necessary and sufficient conditions under which the finite-dimensional approximate Log-HS distance converges to the infinite-dimensional exact Log-HS distance as the dimension goes to infinity. As we prove mathematically, this convergence is non-trivial and requires assumptions beyond those for the approximation of the exact values of the original kernel ($K_1$ in Fig. 1).

For the learning step, with the approximate Log-HS distance, it is then possible to compute efficiently, on large scale datasets, continuous *shift-invariant kernels* ($K_2$ is Fig. 1), such as Gaussian kernels, that approximate the corresponding kernels defined using the exact Log-HS distance.

**Approximate Log-HS inner product**. In contrast to the Log-HS distance, *in the second main theoretical contribution of this work, we show that the infinite-dimensional exact Log-HS inner product can be approximated by its finite-dimensional counterpart only in a very special case*. In fact, in most scenarios, as the dimension approaches infinity, the finite-dimensional approximate Log-HS inner product either diverges or converges to a value that is different from the infinite-dimensional exact Log-HS inner product. For the learning step, this means that kernels defined using the infinite-dimensional Log-HS inner product, such as *linear and polynomial kernels*, are *not* scalable in the same way as the shift-invariant kernels defined using the infinite-dimensional Log-HS distance.

**Covariance operators of low-level features and convolutional features**. While they have been demonstrated to be highly effective for many vision tasks, to the best of our knowledge, covariance descriptors have only been used so far to encode correlations of low-level image features such as colors and gradients, or pre-defined features such as SIFT. This leads us to *convolutional networks* (ConvNets), in which the features are learned by the networks themselves. ConvNets have been shown to scale well to very large datasets, yielding impressive results on many challenging vision tasks recently, including object recognition [47], object localization and detection [19], fine-grained categorization [21], and depth estimation [15], among many others. In this work, apart from covariance operators of low-level and pre-defined image features, we propose to also build covariance operators of convolutional features obtained from ConvNets, specifically the internal representation known as *hypercolumns* [21]. This approach allows us to capture non-linear correlations between convolutional features, further enhancing their expressive power.

**Contributions of this work**. In summary, the novel contributions of our work are the following. *Mathematically and computationally*, we present an approximate formulation for the Log-HS distance between covariance operators that is substantially faster to compute than the exact formulation in [36] and that is scalable to large datasets, while substantially maintaining an effective discriminating capability. Our approximate Log-HS distance formulation is fully justified mathematically, with necessary and sufficient conditions guaranteeing its convergence to the exact Log-HS distance. On the other hand, we show that the exact infinite-dimensional Log-HS inner product can be approximated by its finite-dimensional version in a very limited scenario. Consequently, it cannot be scaled up to large datasets in the same way as the Log-HS distance. *Methodologically*, to our best knowledge, our framework is the first to encode the features of convolutional networks by covariance descriptors, exploiting both the expressiveness of these features and the power of kernel methods and Riemannian geometry. *Empirically*, we apply our approximate Log-HS distance framework to the task of visual object recognition, using altogether twelve challenging, publicly available datatsets, ranging from Fish, Virus, and Texture to Scene recognition. On these datasets, our proposed method compares very favorably with previous state of the art methods in terms of classification accuracy and especially in computational efficiency, demonstrating the competitiveness and potential of our framework.

**Previous and related work.** Infinite-dimensional covariance operators of low-level features have been applied to the task of image classification recently by [36], which formulated the Log-HS metric, and by [20], which used the formulation for Bregman divergences proposed in [57]. While they both work very well, these methods tend to be computationally intensive and not scalable to large datasets. There exists a large literature on large-scale kernel approximation, which focuses on approximating either the feature maps or the kernel matrices [43], [46], [55], but not on the approximation of the covariance operators, as we do in this paper. In [17], the authors essentially consider an approximate version of the affine-invariant distance between covariance operators for image classification. However, the affine-invariant distance generally does *not* lead to positive definite kernels [22] and, as we show in Section 3, this approximation approach is *not* scalable to large datasets.

The original Log-HS distance formulation between positive definite operators in [36] is the infinite-dimensional generalization of the Log-Euclidean distance between SPD matrices in [2]. Recently, the Log-Euclidean metric has been employed to define positive definite kernels, e.g. Gaussian kernels, allowing kernel methods to be applied directly on the manifold of SPD matrices [22], [23], [30]. Thus, the two-layer kernel machine that we describe here, using either the exact Log-HS distance or its approximation, is a generalization of the kernel methods using the Log-Euclidean distance in [22], [23]. In particular, if we set the kernel $K_1$ in Fig. 1 to be the linear kernel, then we recover the framework in [22], [23]. We compare this framework with our current approach empirically in Section 6.

Apart from the affine-invariant Riemannian and Log-Euclidean metrics, one other important family of distance (and distance-like) functions on the set of SPD matrices is that of the Bregman divergences [9], in particular the symmetric Stein divergence [48]. These distance functions do not arise from Riemannian metrics but from the convex structure of the set of SPD matrices and have been shown to work well in applications using nearest neighbor search [11]. In [48], it was shown that, in contrast to the case of the Log-Euclidean distance, the Gaussian kernel defined using the symmetric Stein divergence is positive definite for specific choices of the kernel bandwidth, and thus it might not always be suitable for use in kernel methods, where parameter tuning is often necessary. Kernelized versions of the Bregman divergences and their applications in computer vision have been considered in [20], [57], with strong improvements over the classical Bregman divergences between SPD matrices. To the best of our knowledge, however, there is currently no theoretical result on the positive definiteness of the Gaussian kernel defined using the kernelized symmetric Stein divergence comparable to those given in [48]. We provide empirical comparisons of our current approach with both the symmetric Stein divergence [11] and the kernelized Bregman divergences [20] in Section 6.

We note that, in general, the construction of positive definite kernels using Riemannian metrics, and non-Euclidean metrics generally, is highly non-trivial and is a topic of continuing research interest [18], [48].

The current paper is a significantly extended version of our conference paper [34], with the following novel theoretical, methodological, and empirical contributions, in addition to those in [34]. *Theoretically*, we show that the infinite-dimensional Log-HS inner product between covariance operators in [36] can be approximated by its finite-dimensional counterpart only in a very special case. *Methodologically*, we propose to apply our framework to convolutional features extracted from ConvNet. *Empirically*, we tested our framework using convolutional features on eight datasets, obtaining very favorable results compared to the state of the art.

**Organization.** We give an overview of the Riemannian distances between finite-dimensional covariance matrices and their generalizations to infinite-dimensional covariance operators in Section 2. The core of the paper is contained in Section 3, in which we present our approximate Log-HS distance formulation, using two methods for computing approximate feature maps and covariance operators. Section 4 contains our analysis of the approximate Log-HS inner product, in particular its non-convergence to the exact infinite-dimensional Log-HS inner product. Section 5 describes two feature representations, namely low-level and convolutional features, that are used in the experiments. Empirical results on the task of visual object recognition, using twelve different datasets, are reported in Section 6. Proofs for all mathematical results in the paper, along with further technical information, are given in the Appendices.

## 2 DISTANCES BETWEEN COVARIANCE MATRICES AND COVARIANCE OPERATORS

Let $n \in \mathbb{N}$ be fixed. Given two $n \times n$ covariance matrices $A$ and $B$, a straightforward distance between them is the Euclidean distance

$$d_E(A, B) = ||A - B||_F, \quad (1)$$

where $|| \ ||$ denotes the Frobenius norm, which for $A = (a_{ij})_{i,j=1}^n$ is given by $||A||_F^2 = \sum_{i,j=1}^n a_{ij}^2$. This norm is induced by the Frobenius inner product $\langle A, B \rangle_F = \text{tr}(A^T B)$. Clearly, both the distance $||A - B||_F$ and the inner product $\langle A, B \rangle_F$, which are defined solely in terms of the entries of $A$ and $B$, do *not* reflect any structure in $A$ and $B$, including in particular the fact that they are both elements of the convex cone of positive semi-definite matrices. In order to do so, the following distances, which are based on the intrinsic geometry of the set of SPD matrices, are commonly employed in the literature.

## 2.1 Riemannian distances between SPD matrices

Covariance matrices of size $n \times n$, properly regularized if necessary, are instances of the set $\mathrm{Sym}^{++}(n)$ of SPD matrices, which can be endowed with a Riemannian metric structure, forming a finite-dimensional Riemannian manifold, see *e.g.* [3], [41]. The most commonly encountered Riemannian metric on $\mathrm{Sym}^{++}(n)$ is the *affine-invariant metric*, in which the geodesic distance between two SPD matrices $A$ and $B$ is given by

$$d_{\mathrm{aiE}}(A, B) = ||\log(A^{-1/2}BA^{-1/2})||_F. \tag{2}$$

In Eq.(2), log denotes the principal matrix logarithm, which for an SPD matrix $A$ is defined by $\log(A) = U \log(D) U^T$, where $A = UDU^T$ is the spectral decomposition for $A$. From a practical perspective, the distance (2) tends to be computationally intensive for large scale datasets. This motivated the development of the *Log-Euclidean metric* framework of [2], in which the geodesic distance between $A$ and $B$ is given by

$$d_{\mathrm{logE}}(A, B) = ||\log(A) - \log(B)||_F. \tag{3}$$

The Log-Euclidean distance given in (3) is faster to compute than the affine-invariant distance (2), especially for computing all pairwise distances on a large set of SPD matrices. Furthermore, the Log-Euclidean metric can be used to define many positive definite kernels, such as the Gaussian kernel, which is not possible using the affine-invariant metric [22].

**Log-Euclidean distance versus Euclidean distance.** While these two distances are closely linked to each other, *we wish to emphasize that the Log-Euclidean metric is itself a Riemannian metric on* $\mathrm{Sym}^{++}(n)$, *a so-called bi-invariant metric*. This metric arises from the Lie group structure of $\mathrm{Sym}^{++}(n)$ corresponding to the commutative group operation $\odot : \mathrm{Sym}^{++}(n) \times \mathrm{Sym}^{++}(n) \to \mathrm{Sym}^{++}(n)$, defined by

$$A \odot B = \exp(\log(A) + \log(B)). \tag{4}$$

This Lie group structure was introduced in [2], where the corresponding geodesic curves and the geodesic distance (3) are derived. In particular, for two *commuting* matrices $A$ and $B$, the Log-Euclidean and affine-invariant distances are identical, i.e. $d_{\mathrm{logE}}(A, B) = d_{\mathrm{aiE}}(A, B)$. Along with the operation $\odot$, consider the operation $\circledast : \mathbb{R} \times \mathrm{Sym}^{++}(n) \to \mathrm{Sym}^{++}(n)$, defined by

$$\lambda \circledast A = \exp(\lambda \log(A)) = A^{\lambda}. \tag{5}$$

The algebraic structure $(\mathrm{Sym}^{++}(n), \odot, \circledast)$ is then a vector space, with $\odot$ acting as vector addition and $\circledast$ acting as scalar multiplication. The vector space $(\mathrm{Sym}^{++}(n), \odot, \circledast)$ can be endowed with the *Log-Euclidean inner product*, defined by [30]

$$\langle A, B \rangle_{\mathrm{logE}} = \langle \log(A), \log(B) \rangle_F = \mathrm{tr}[\log(A) \log(B)]. \tag{6}$$

The algebraic structure $(\mathrm{Sym}^{++}(n), \odot, \circledast, \langle \ , \ \rangle_{\mathrm{logE}})$ is then an inner product space and the Log-Euclidean distance $||\log(A) - \log(B)||_F$ is precisely the Hilbert distance induced by this inner product. Furthermore, let $(\mathrm{Sym}(n), +, \cdot)$ denote the vector space of $n \times n$ symmetric matrices under the standard matrix addition $+$ and scalar multiplication $\cdot$ operations, then the map

$$\log : (\mathrm{Sym}^{++}(n), \odot, \circledast, \langle \ , \ \rangle_{\mathrm{logE}}) \to (\mathrm{Sym}(n), +, \cdot, \langle \ , \ \rangle_F), \tag{7}$$
$$A \to \log(A),$$

is an isometry. Thus $(\mathrm{Sym}^{++}(n), \odot, \circledast, \langle \ , \ \rangle_{\mathrm{logE}})$ is isometric to the Euclidean space $(\mathrm{Sym}(n), +, \cdot, \langle \ , \ \rangle_F)$. Consequently, under the Log-Euclidean metric, $\mathrm{Sym}^{++}(n)$ is a Riemannian manifold with zero sectional curvature.

The fact that the Log-Euclidean distance is a Hilbert distance is precisely the reason why it can be used to define many positive definite kernels, such as the Gaussian kernels. *However*, the *isometry* given in (7) does *not* imply that the Log-Euclidean distance behaves in the same way as the Euclidean distance $||A - B||_F$. In contrast to the Euclidean distance $||A - B||_F$, which does not encode any structure of $A$ and $B$, the Log-Euclidean distance $||\log(A) - \log(B)||$ encodes the positivity of $A$ and $B$ through the principal logarithm log (in fact, if $A$ has a negative real eigenvalue, then $\log(A)$ would *not* even be defined). Thus, even though $\mathrm{Sym}^{++}(n)$ has zero curvature under the Log-Euclidean metric, the geodesic distance (3) nevertheless captures the intrinsic geometry of $\mathrm{Sym}^{++}(n)$ better than the Euclidean distance $||A - B||_F$, which is extrinsic to $\mathrm{Sym}^{++}(n)$. This has also been consistently demonstrated empirically, see e.g [2], [22].

**Covariance matrices of features.** The SPD matrices considered in the current work are covariance matrices of features extracted from input data. Specifically, let $\mathcal{X} \subset \mathbb{R}^n$. Let $\mathbf{x} = [x_1, \dots, x_m]$ be a data matrix sampled from $\mathcal{X}$, where $m$ is the number of observations. In the setting of the current work, there is one such matrix for each image, namely the matrix of low-level features sampled at (a subset of) the pixels in the image. Each image is then represented by the $n \times n$ covariance matrix

$$C_{\mathbf{x}} = \frac{1}{m} \mathbf{x} J_m \mathbf{x}^T : \mathbb{R}^n \to \mathbb{R}^n, \tag{8}$$

where $J_m$ is the centering matrix, defined by $J_m = I_m - \frac{1}{m} \mathbf{1}_m \mathbf{1}_m^T$ with $\mathbf{1}_m = (1, \dots, 1)^T \in \mathbb{R}^m$. In practice, $C_{\mathbf{x}}$ is generally only positive semi-definite and thus to apply the Riemannian structure of $\mathrm{Sym}^{++}(n)$, it is often necessary to consider the regularized version $(C_{\mathbf{x}} + \gamma I_n)$ for some $\gamma > 0$. For two covariance matrices $C_{\mathbf{x}}$ and $C_{\mathbf{y}}$, we therefore consider the distance between the regularized versions $(C_{\mathbf{x}} + \gamma I)$ and $(C_{\mathbf{y}} + \mu I)$, given by

$$d_{\mathrm{logE}} = ||\log(C_{\mathbf{x}} + \gamma I_n) - \log(C_{\mathbf{y}} + \mu I_n)||_F, \tag{9}$$

for some regularization parameters $\gamma > 0, \mu > 0$.

## 2.2 Infinite-dimensional covariance operators

The covariance matrix $C_{\mathbf{x}}$ only measures the *linear* correlations between the features in the input data. A powerful method to capture *non-linear* input correlations is by (i) first mapping the original input features into a high dimensional feature space $\mathcal{H}$, using an implicit nonlinear feature map induced by a positive definite kernel; (ii) then computing the covariance operators in the feature space $\mathcal{H}$.

Specifically, let $\mathcal{X}$ be an arbitrary non-empty set. Let $\mathbf{x} = [x_1, \dots, x_m]$ be a data matrix sampled from $\mathcal{X}$, where $m \in \mathbb{N}$ is the number of observations. Let $K$ be a positive definite kernel on $\mathcal{X} \times \mathcal{X}$ and $\mathcal{H}_K$ its induced reproducing kernel Hilbert space (RKHS). Let $\mathcal{H}$ be any feature space for $K$, which we assume to be a separable Hilbert space, with the corresponding feature map $\Phi : \mathcal{X} \to \mathcal{H}$, so that $K(x, y) = \langle \Phi(x), \Phi(y) \rangle_{\mathcal{H}}$ for all pairs $(x, y) \in \mathcal{X} \times \mathcal{X}$. For concreteness, we can identify $\mathcal{H}$ with the RKHS $\mathcal{H}_K$, or with the space of square summable sequences $\ell^2 = \{(a_k)_{k \in \mathbb{N}} : \sum_{k=1}^{\infty} |a_k|^2 < \infty\}$. The feature map $\Phi$ gives the (potentially infinite) mapped data matrix

$$\Phi(\mathbf{x}) = [\Phi(x_1), \dots, \Phi(x_m)] \tag{10}$$

of size $\dim(\mathcal{H}) \times m$, with the $j$th column being $\Phi(x_j)$, in the feature space $\mathcal{H}$. Formally, $\Phi(\mathbf{x})$ is a bounded linear operator $\Phi(\mathbf{x}) : \mathbb{R}^m \to \mathcal{H}$, defined by

$$\Phi(\mathbf{x})\mathbf{b} = \sum_{j=1}^{m} b_j \Phi(x_j), \quad \mathbf{b} \in \mathbb{R}^m. \tag{11}$$

Let $\Phi(\mathbf{x})^* : \mathcal{H} \to \mathbb{R}^m$ denote the adjoint operator of $\Phi(\mathbf{x})$, which is simply the transpose $\Phi(\mathbf{x})^T$ if $\dim(\mathcal{H}) < \infty$. The corresponding covariance operator for $\Phi(\mathbf{x})$ is then defined to be

$$C_{\Phi(\mathbf{x})} = \frac{1}{m} \Phi(\mathbf{x}) J_m \Phi(\mathbf{x})^* : \mathcal{H} \to \mathcal{H}, \tag{12}$$

which can be considered as a (potentially infinite) matrix of size $\dim(\mathcal{H}) \times \dim(\mathcal{H})$. *The view of $\Phi(\mathbf{x})$ and $C_{\Phi(\mathbf{x})}$ as infinite matrices emphasizes their connection with the finite-dimensional setting, given by Eq. (8).* In particular, if $\mathcal{X} = \mathbb{R}^n$ and $K(x,y) = \langle x, y \rangle$, then $C_{\Phi(\mathbf{x})} = C_{\mathbf{x}}$ as in Eq. (8). The operator $C_{\Phi(\mathbf{x})}$ is self-adjoint, positive, and with rank at most $m - 1$. For a more formal description of $C_{\Phi(\mathbf{x})}$, we refer to [35].

**Hilbert-Schmidt distance**. Given two infinite-dimensional covariance operators $C_{\Phi(\mathbf{x})}$ and $C_{\Phi(\mathbf{y})}$ on $\mathcal{H}$, a natural distance between them is the Hilbert-Schmidt distance, which is the generalization of the Frobenius distance $|| \ ||_F$ to the infinite dimensional setting. Let $A : \mathcal{H} \to \mathcal{H}$ be a bounded linear operator and $A^*$ be its adjoint operator. We recall that $A : \mathcal{H} \to \mathcal{H}$ is said to be a Hilbert-Schmidt operator, denoted by $A \in \mathrm{HS}(\mathcal{H})$, if

$$||A||_{\mathrm{HS}}^2 = \mathrm{tr}(A^*A) = \sum_{k=1}^{\infty} \lambda_k(A^*A) < \infty, \qquad (13)$$

where $|| \ ||_{\mathrm{HS}}$ denotes the Hilbert-Schmidt norm, the infinite-dimensional generalization of the Frobenius norm, and $\{\lambda_k(A^*A)\}_{k=1}^{\infty}$ denotes the set of eigenvalues of $A^*A$. Given two covariance operators $C_{\Phi(\mathbf{x})}$ and $C_{\Phi(\mathbf{y})}$, the Hilbert-Schmidt distance between them is then defined by

$$d_{\mathrm{HS}}[C_{\Phi(\mathbf{x})}, C_{\Phi(\mathbf{y})}] = ||C_{\Phi(\mathbf{x})} - C_{\Phi(\mathbf{y})}||_{\mathrm{HS}}. \qquad (14)$$

This distance admits a closed form in terms of Gram matrices (the explicit expression is given in Appendix B). However, as with the Frobenius distance $|| \ ||_F$, the Hilbert-Schmidt distance does not reflect the positivity of $C_{\Phi(\mathbf{x})}$ and $C_{\Phi(\mathbf{y})}$. In order to do so, we consider the following generalizations of the affine-invariant and Log-Euclidean metrics.

**Infinite-dimensional affine-invariant distance**. As in the finite-dimensional case, for the generalizations of the affine-invariant and Log-Euclidean metrics, we need to consider the regularized covariance operator $(C_{\Phi(\mathbf{x})} + \gamma I_{\mathcal{H}})$, $\gamma > 0$, which lies on the infinite-dimensional manifold $\Sigma(\mathcal{H})$ of positive definite operators on $\mathcal{H}$. By the formulation of [28], [33], the infinite-dimensional affine-invariant distance $d_{\mathrm{aiHS}}$ between $(C_{\Phi(\mathbf{x})} + \gamma I)$ and $(C_{\Phi(\mathbf{y})} + \mu I)$ is given by

$$d_{\mathrm{aiHS}}[(C_{\Phi(\mathbf{x})} + \gamma I), (C_{\Phi(\mathbf{y})} + \mu I)] \qquad (15)$$
$$= || \log[(C_{\Phi(\mathbf{x})} + \gamma I)^{-1/2}(C_{\Phi(\mathbf{y})} + \mu I)(C_{\Phi(\mathbf{x})} + \gamma I)^{-1/2}]||_{\mathrm{eHS}},$$

with the extended Hilbert-Schmidt norm $|| \ ||_{\mathrm{eHS}}$ given by $||A + \gamma I||_{\mathrm{eHS}}^2 = ||A||_{\mathrm{HS}}^2 + \gamma^2$. The explicit expression for this distance in terms of Gram matrices is given in [33].

**Log-Hilbert-Schmidt distance**. The generalization of the Log-Euclidean metric to the infinite-dimensional manifold $\Sigma(\mathcal{H})$ has recently been formulated by [36]. In this metric, termed *Log-Hilbert-Schmidt metric*, or *Log-HS* for short, the distance $d_{\mathrm{logHS}}[(C_{\Phi(\mathbf{x})} + \gamma I_{\mathcal{H}}), (C_{\Phi(\mathbf{y})} + \mu I_{\mathcal{H}})]$ between $(C_{\Phi(\mathbf{x})} + \gamma I_{\mathcal{H}})$ and $(C_{\Phi(\mathbf{y})} + \mu I_{\mathcal{H}})$ is given by

$$d_{\mathrm{logHS}}[(C_{\Phi(\mathbf{x})} + \gamma I_{\mathcal{H}}), (C_{\Phi(\mathbf{y})} + \mu I_{\mathcal{H}})] \qquad (16)$$
$$= || \log(C_{\Phi(\mathbf{x})} + \gamma I_{\mathcal{H}}) - \log(C_{\Phi(\mathbf{y})} + \mu I_{\mathcal{H}})||_{\mathrm{eHS}},$$

which has a closed form in terms of the corresponding Gram matrices (we refer to [36] for the explicit expression).

**Regularization**. The form of regularization $(A + \gamma I)$, $\gamma > 0$, which is often *empirically* necessary to ensure positive definiteness in the case $\dim(\mathcal{H}) < \infty$, is always necessary, both *theoretically* and *empirically*, in the case $\dim(\mathcal{H}) = \infty$, since in this case $\log(A)$, with $A$ being a compact operator, is always unbounded (see [28], [33], [36]).

As shown in [36], both of the operations $\odot$ and $\circledast$, defined in (4) and (5), respectively, admit generalizations on the manifold $\Sigma(\mathcal{H})$, making $(\Sigma(\mathcal{H}), \odot, \circledast)$ a vector space. Furthermore, one can define the *Log-Hilbert-Schmidt inner product* on $\Sigma(\mathcal{H})$, which, in the current setting of covariance operators, takes the form

$$\langle (C_{\Phi(\mathbf{x})} + \gamma I), (C_{\Phi(\mathbf{y})} + \mu I) \rangle_{\mathrm{logHS}}$$
$$= \langle \log(C_{\Phi(\mathbf{x})} + \gamma I), \log(C_{\Phi(\mathbf{y})} + \mu I) \rangle_{\mathrm{eHS}}. \qquad (17)$$

The Log-Hilbert-Schmidt distance is then precisely the Hilbert distance corresponding to this inner product.

As in the finite-dimensional case, *two key advantages of the Log-HS distance $d_{\mathrm{logHS}}$ over the affine-invariant distance $d_{\mathrm{aiHS}}$* are: (i) the $d_{\mathrm{logHS}}$ distance is faster to compute than the $d_{\mathrm{aiHS}}$ distance; (ii) since $d_{\mathrm{logHS}}$ is a Hilbert distance, it can be used to to define many commonly used positive definite kernels, such as the Gaussian kernel, which is not the case with the $d_{\mathrm{aiHS}}$ distance. These advantages are fully exploited in the current work.

In the next section, we describe how to approximate Formula (16) to compute the pairwise distances on a set of data matrices $\{\mathbf{x}_i\}_{i=1}^N$ when $N$ and $m$ are large.

# 3 APPROXIMATE LOG-HS DISTANCE

While kernel methods are powerful in learning non-linear structures in data, they tend not to scale well, in terms of computational complexity, to large datasets, which are common in vision problems such as object recognition and fine-grained categorization. In the typical kernel learning setting, the feature map $\Phi$ is high-dimensional (and often infinite-dimensional, as in the case of the Gaussian kernel) and thus is only used *implicitly*. Instead, exact kernel methods carry out computations using Gram matrices and thus their computational complexities depend on the sizes of the Gram matrices, which become very large for large datasets.

A commonly used approach that has emerged recently to reduce the computational cost of kernel methods is to compute an *explicit approximate feature map* $\hat{\Phi}_D : \mathcal{X} \to \mathbb{R}^D$, where $D$ is finite and $D << \dim(\mathcal{H})$, so that

$$\langle \hat{\Phi}_D(x), \hat{\Phi}_D(y) \rangle_{\mathbb{R}^D} = \hat{K}_D(x,y) \approx K(x,y), \quad \text{with} \qquad (18)$$

$$\lim_{D \to \infty} \hat{K}_D(x,y) = K(x,y), \qquad (19)$$

$\forall (x,y) \in \mathcal{X} \times \mathcal{X}$. The approximate feature map $\hat{\Phi}_D$ is then used directly in the learning algorithms instead of the Gram matrices.

In our setting, we use the approximate feature maps to compute the corresponding finite-dimensional approximate covariance operators. The Log-Euclidean distance between the approximate operators is then used as the approximate version of Log-HS distance between the infinite-dimensional covariance operators. *Thus we are not interested in the approximate kernel values $\hat{K}_D(x,y)$ per se, but the approximate covariance operators and the corresponding approximate Log-HS distance. The mathematical justification for the latter goes beyond that for kernel approximation and is the first main theoretical contribution of this work.* In fact, as we show in Theorems 1 and 2 below, Eq. (19), which guarantees the convergence of the approximate kernel value to the true kernel value, is *not sufficient* to guarantee the convergence of the approximate Log-HS distance to the true Log-HS distance as $D \to \infty$. This convergence is non-trivial and requires further assumptions, which are practically realizable, as we explain below.

**Approximate covariance operator and approximate Log-HS distance**. With the approximate feature map $\hat{\Phi}_D$, we have the corresponding data matrix $\hat{\Phi}_D(\mathbf{x}) = [\hat{\Phi}_D(x_1), \dots, \hat{\Phi}_D(x_m)]$ of size $D \times m$, and the approximate covariance operator has the form

$$C_{\hat{\Phi}_D(\mathbf{x})} = \frac{1}{m} \hat{\Phi}_D(\mathbf{x}) J_m \hat{\Phi}_D(\mathbf{x})^T : \mathbb{R}^D \to \mathbb{R}^D, \qquad (20)$$

which is a matrix of size $D \times D$, instead of the potentially infinite matrix $C_{\Phi(\mathbf{x})}$ of size $\dim(\mathcal{H}) \times \dim(\mathcal{H})$.

We then consider the following as an approximate version of the Log-HS distance given in Formula (16):

$$\left\| \log\left(C_{\hat{\Phi}_D(\mathbf{x})} + \gamma I_D\right) - \log\left(C_{\hat{\Phi}_D(\mathbf{y})} + \mu I_D\right) \right\|_F. \quad (21)$$

**Key theoretical question I**. We need to determine whether Formula (21) is truly a finite-dimensional approximation of Formula (16), in the sense that

$$\lim_{D \to \infty} \left\| \log(C_{\hat{\Phi}_D(\mathbf{x})} + \gamma I_D) - \log(C_{\hat{\Phi}_D(\mathbf{y})} + \mu I_D) \right\|_F$$
$$= || \log(C_{\Phi(\mathbf{x})} + \gamma I_{\mathcal{H}}) - \log(C_{\Phi(\mathbf{y})} + \mu I_{\mathcal{H}})||_{\text{eHS}}. \quad (22)$$

The answer to this question is the *first main mathematical contribution* of the current paper. It turns out that in general, this is *not* possible. This is because the infinite-dimensional Log-HS distance is generally *not* obtainable as a limit of the finite-dimensional Log-Euclidean distance as the dimension approaches infinity [36]. More precisely, we have the following result.

**Theorem 1.** *Assume that* $\gamma \neq \mu$, $\gamma > 0$, $\mu > 0$. *Then*

$$\lim_{D \to \infty} \left\| \log(C_{\hat{\Phi}_D(\mathbf{x})} + \gamma I_D) - \log(C_{\hat{\Phi}_D(\mathbf{y})} + \mu I_D) \right\|_F = \infty.$$

The infinite limit in Theorem 1 stands in sharp contrast to that of Eq. (19) on the approximability of the kernel value $K(x, y)$ itself, which is satisfied by both approximation schemes based on Fourier features presented below.

In practice, however, it is reasonable to assume that we can use the same regularization parameter for both $C_{\hat{\Phi}_D(\mathbf{x})}$ and $C_{\hat{\Phi}_D(\mathbf{y})}$, that is to set $\gamma = \mu$. In this setting, we obtain the necessary convergence, as follows.

**Theorem 2.** *Assume that* $\dim(\mathcal{H}) = \infty$. *Assume that* $\gamma = \mu > 0$. *Then*

$$\lim_{D \to \infty} \left\| \log(C_{\hat{\Phi}_D(\mathbf{x})} + \gamma I_D) - \log(C_{\hat{\Phi}_D(\mathbf{y})} + \gamma I_D) \right\|_F$$
$$= || \log(C_{\Phi(\mathbf{x})} + \gamma I_{\mathcal{H}}) - \log(C_{\Phi(\mathbf{y})} + \gamma I_{\mathcal{H}})||_{\text{eHS}}. \quad (23)$$

In light of Theorems 1 and 2, subsequently we employ the same regularization parameter $\gamma > 0$ to compute approximate Log-HS distances between all regularized operators $(C_{\hat{\Phi}_D(\mathbf{x})} + \gamma I_D)$. In this work, we focus on shift-invariant kernels and employ two methods for computing the corresponding approximate feature map $\hat{\Phi}_D$, namely Random Fourier features [43] and Quasi-random Fourier features [55], presented in the following section.

### 3.1 Fourier feature maps

**Random Fourier feature maps**. This is the approach in [43] for computing approximate feature maps of shift-invariant kernels. Let $K : \mathbb{R}^n \times \mathbb{R}^n \to \mathbb{R}$ be a kernel of the form $K(x, y) = k(x - y)$ for some positive definite function $k$ on $\mathbb{R}^n$. By Bochner's Theorem [44], there is a finite positive measure $\rho$ on $\mathbb{R}^n$ such that

$$k(x - y) = \int_{\mathbb{R}^n} e^{-i\langle \omega, x-y \rangle} d\rho(\omega) \quad (24)$$
$$= \int_{\mathbb{R}^n} \phi_\omega(x)\overline{\phi_\omega(y)} d\rho(\omega), \text{ where } \phi_\omega(x) = e^{-i\langle \omega, x \rangle}.$$

Without loss of generality, we can assume that $\rho$ is a probability measure on $\mathbb{R}^n$, so that $K(x, y) = \mathbb{E}_{\omega \sim \rho}[\phi_\omega(x)\overline{\phi_\omega(y)}]$. By symmetry, $K(x, y) = \frac{1}{2}[K(x, y) + K(y, x)]$, so that by the relation $\frac{1}{2}[e^{-i\langle \omega, x-y \rangle} + e^{i\langle \omega, x-y \rangle}] = \cos(\langle \omega, x - y \rangle)$ we have

$$K(x, y) = \int_{\mathbb{R}^n} \cos(\langle \omega, x - y \rangle) d\rho(\omega). \quad (25)$$

To approximate $K(x, y)$, we can sample $D$ points $\{\omega_j\}_{j=1}^D$ from the distribution $\rho$ and compute the empirical version

$$\hat{K}_D(x, y) = \frac{1}{D} \sum_{j=1}^D \cos(\langle \omega_j, x - y \rangle) \xrightarrow{D \to \infty} K(x, y) \quad (26)$$

almost surely by the law of large numbers. Let $W = (\omega_1, \ldots, \omega_D)$ be an $n \times D$ matrix with each column $\omega_j \in \mathbb{R}^n$ randomly sampled according to $\rho$. Using the cosine addition formula $\cos\langle \omega_j, x - y \rangle = \cos\langle \omega_j, x \rangle \cos\langle \omega_j, y \rangle + \sin\langle \omega_j, x \rangle \sin\langle \omega_j, y \rangle$, we define

$$\cos(W^T x) = (\cos\langle \omega_j, x \rangle)_{j=1}^D, \quad (27)$$
$$\sin(W^T x) = (\sin\langle \omega_j, x \rangle)_{j=1}^D. \quad (28)$$

The desired approximate feature map is the concatenation

$$\hat{\Phi}_D(x) = \frac{1}{\sqrt{D}}(\cos(W^T x); \sin(W^T x)) \in \mathbb{R}^{2D}, \quad (29)$$

with $\langle \hat{\Phi}_D(x), \hat{\Phi}_D(y) \rangle = \hat{K}_D(x, y)$. In the case of the Gaussian kernel (used in the experiments in Sec. 6) $K(x, y) = e^{-\frac{||x-y||^2}{\sigma^2}}$, $\rho$ is a Gaussian probability measure on $\mathbb{R}^n$, with density

$$\rho(\omega) = \frac{(\sigma\sqrt{\pi})^n}{(2\pi)^n} e^{-\frac{\sigma^2 ||\omega||^2}{4}} \sim \mathcal{N}\left(0, \frac{2}{\sigma^2} I_n\right). \quad (30)$$

**Quasi-random Fourier feature maps**. The Random Fourier feature maps above arise from the Monte-Carlo approximation of the kernel $K$ expressed as the integral in Eq. (24), using a *random* set of points $\omega_j$'s sampled according to the distribution $\rho$. An alternative approach, proposed by [55], employs Quasi-Monte Carlo integration [13], in which the $\omega_j$'s are *deterministic* points arising from a *low-discrepancy* sequence in $[0, 1]^n$. Thus the obvious advantage of this approach is that it does not require random sampling. However, compared with the random Fourier feature approach, it requires further assumptions on the probability distribution $\rho$ to be applicable. We describe this approach in detail in Appendix C.

### 3.2 New positive definite kernels using approximate Log-HS distances

In our framework, starting with a shift-invariant kernel $K_1$ in Fig. 1, we compute the approximate feature map $\hat{\Phi}_D(x)$ using the methods in Section 3.1 and the corresponding approximate covariance operators using Eq. (20). The approximate Log-HS distances between these approximate covariance operators are then computed using Eq. (21). With the approximate Log-HS distances, we can define a new positive definite kernel ($K_2$ in Fig. 1), for example

$$\exp(-|| \log(C_{\hat{\Phi}_D(\mathbf{x})} + \gamma I_D) - \log(C_{\hat{\Phi}_D(\mathbf{y})} + \gamma I_D)||_F^p/\sigma^2), \quad (31)$$

for $0 < p \leq 2$, with $p = 2$ giving the Gaussian kernel and $p = 1$ giving the Laplacian kernel. This new kernel can then be used in a classifier, e.g. SVM. The complete pipeline for our framework is summarized in Algorithm 1.

### 3.3 Computational complexity

We present here the computational complexity analysis of the proposed approximation in Eq. (21) as well as the comparison with the approximate affine-invariant distance, which is essentially carried out in [17], according to the formula

$$\left\| \log[(C_{\hat{\Phi}_D(\mathbf{x})} + \gamma I)^{-1/2}(C_{\hat{\Phi}_D(\mathbf{y})} + \mu I)(C_{\hat{\Phi}_D(\mathbf{x})} + \gamma I)^{-1/2}] \right\|_F. \quad (32)$$

The main computational cost in Eq. (21) is the SVD for $(C_{\tilde{\Phi}_D(\mathbf{x})} + \gamma I)$ and $(C_{\tilde{\Phi}_D(\mathbf{y})} + \mu I)$, which takes time $O(D^3)$. At first glance, the computational complexity for the approximate affine-invariant distance in Eq. (32), which consists of a matrix square root and inversion, two matrix multiplications and an SVD, is also $O(D^3)$. However, computationally, the key difference between Eq. (21) and Eq. (32) is that in Eq. (21), $(C_{\tilde{\Phi}_D(\mathbf{x})} + \gamma I)$ and $(C_{\tilde{\Phi}_D(\mathbf{y})} + \mu I)$ are *uncoupled*, whereas in Eq. (32), they are *coupled*. Thus if we have $N$ data matrices $\{\mathbf{x}_1, \ldots, \mathbf{x}_N\}$, to compute their pairwise approximate Log-HS distances using Eq. (21), we need to compute an SVD for each $(C_{\tilde{\Phi}_D(\mathbf{x}_i)} + \gamma I)$, with time complexity $O(ND^3)$. On the other hand, to compute their pairwise approximate affine-invariant distances using Eq. (32), we need to compute an SVD for each *pair* $(C_{\tilde{\Phi}_D(\mathbf{x}_i)} + \gamma I)$, $(C_{\tilde{\Phi}_D(\mathbf{x}_j)} + \gamma I)$, with time complexity $O(N^2 D^3)$. Thus the approximation of the Log-HS distance is $O(N)$ times faster than the approximation of the affine-invariance distance.

We also note that for $N$ pairs of data matrices, the computational complexity of the exact Log-HS formulation [36] and the RKHS Bregman divergences [20] is of order $O(N^2 m^3)$. Thus for $D < m$ and $N$ large, the approximate Log-HS formulation will be much more efficient to compute than both the exact Log-HS and the RKHS Bregman divergences (see the actual running time comparison between the approximate and exact Log-HS formulations in the experiments below).

---

**Input**: Set of images.
**Output**: Kernel matrix (used as input to a classifier, e.g. SVM).
**Parameters**: Kernels $K_1$, $K_2$, regularization parameters $\gamma = \mu > 0$, approximate feature dimension $D$.
**Procedure**:

1) For each image, extract a data matrix $\mathbf{x} = [x_1, \ldots, x_m]$ of low-level features from $m$ pixels.

2) For each image, compute the approximate feature maps $\hat{\Phi}_D(x_i)$, $1 \leq i \leq m$, associated to kernel $K_1$, according to Eq. (29), and the corresponding approximate covariance operator $C_{\tilde{\Phi}_D(\mathbf{x})}$, according to Eq. (20).

3) For each pair of images, compute the approximate Log-HS distance between the corresponding covariance operators, according to Eq. (21).

4) Using kernel $K_2$, compute a kernel matrix using the above approximate Log-HS distance, e.g. according to Eq. (31).

Algorithm 1: Summary of the proposed method.

---

**Further comparison with [17]**. In [17], the authors proposed two methods: (i) Nearest Neighbor using effectively the approximate affine-invariant distance given in Eq. (32) with $\gamma = \mu = 0$ and (ii) the CDL algorithm [53] using the representation $\log(C_{\tilde{\Phi}_D(\mathbf{x})})$. Both of these methods require the assumption that $C_{\tilde{\Phi}_D(\mathbf{x})}$ is positive definite, which is *never* guaranteed. In fact, when $D > m$, $C_{\tilde{\Phi}_D}(\mathbf{x})$ is always rank-deficient and neither its inverse nor log can be computed. Thus neither the CDL nor the approximate affine-invariant distance can be used. Theoretically, since it does not employ any regularization, the approximate affine-invariant distance in [17] will *not* approach the exact affine-invariant distance ( [28], [33]) for large $D$, which *always* requires regularization (see Sec. 2.2).

## 4 APPROXIMATE LOG-HS INNER PRODUCT

We have shown in Section 3, via Theorems 1 and 2, that the infinite-dimensional Log-HS distance $||\log(C_{\Phi(\mathbf{x})} + \gamma I_{\mathcal{H}}) -$

$\log(C_{\Phi(\mathbf{y})} + \mu I_{\mathcal{H}})||_{\text{eHS}}$ can be approximated by the finite-dimensional Log-Euclidean distance $||\log(C_{\tilde{\Phi}_D(\mathbf{x})} + \gamma I_D) - \log(C_{\tilde{\Phi}_D(\mathbf{y})} + \mu I_D)||_F$ if and only if $\gamma = \mu > 0$. This in turns implies that, for $\gamma = \mu > 0$, the Gaussian kernel $\exp(-||\log(C_{\Phi(\mathbf{x})} + \gamma I_{\mathcal{H}}) - \log(C_{\Phi(\mathbf{y})} + \gamma I_{\mathcal{H}})||^2_{\text{eHS}}/\sigma^2)$, which is defined on the infinite-dimensional manifold $\Sigma(\mathcal{H})$, can be approximated by the Gaussian kernel $\exp(-||\log(C_{\tilde{\Phi}_D(\mathbf{x})} + \gamma I_D) - \log(C_{\tilde{\Phi}_D(\mathbf{y})} + \gamma I_D)||^2_F/\sigma^2)$, which is defined on the finite-dimensional manifold $\text{Sym}^{++}(D)$. The same statement holds true when the Gaussian kernel is replaced by any continuous positive definite shift-invariant kernels defined on $\Sigma(\mathcal{H})$.

In this section, we investigate the approximability of the infinite-dimensional Log-HS inner product $\langle \rangle_{\text{logHS}}$, as defined by Eq. (17), by the finite-dimensional Log-Euclidean inner product $\langle \rangle_{\text{logE}}$, as defined by Eq. (6).

**Key theoretical question II**. Assuming that we have the kernel approximation given by Eq. (18) and Eq. (19), we need to determine the conditions under which

$$\lim_{D \to \infty} \left\langle \log(C_{\tilde{\Phi}_D(\mathbf{x})} + \gamma I_D), \log(C_{\tilde{\Phi}_D(\mathbf{y})} + \mu I_D) \right\rangle_F = \left\langle \log(C_{\Phi(\mathbf{x})} + \gamma I_{\mathcal{H}}), \log(C_{\Phi(\mathbf{y})} + \mu I_{\mathcal{H}}) \right\rangle_{\text{eHS}}. \quad (33)$$

The answer to this question is the *second main mathematical contribution* of the current paper. It turns out that this is only true in the special case $\gamma = \mu = 1$. For all other values of $\gamma > 0, \mu > 0$, the left hand side of Eq. (33) either approaches infinity or a finite limit that is *not* equal to the right hand side. More precisely, we have the following result.

**Theorem 3.** *Assume that* $\dim(\mathcal{H}) = \infty$. *Let* $\gamma > 0$, $\mu > 0$. *There are four possible cases, as follows.*

*1) If* $\gamma = \mu = 1$, *then*

$$\lim_{D \to \infty} \left\langle \log(C_{\tilde{\Phi}_D(\mathbf{x})} + I_D), \log(C_{\tilde{\Phi}_D(\mathbf{y})} + I_D) \right\rangle_F = \left\langle \log(C_{\Phi(\mathbf{x})} + I_{\mathcal{H}}), \log(C_{\Phi(\mathbf{y})} + I_{\mathcal{H}}) \right\rangle_{\text{eHS}}. \quad (34)$$

*2) If* $\gamma = 1$, $\mu \neq 1$, *then*

$$\lim_{D \to \infty} \left\langle \log(C_{\tilde{\Phi}_D(\mathbf{x})} + I_D), \log(C_{\tilde{\Phi}_D(\mathbf{y})} + \mu I_D) \right\rangle_F = \left\langle \log(C_{\Phi(\mathbf{x})} + I_{\mathcal{H}}), \log(C_{\Phi(\mathbf{y})} + \mu I_{\mathcal{H}}) \right\rangle_{\text{eHS}} + (\log \mu) \text{tr} \left[ \log \left( \frac{1}{m} J_m K[\mathbf{x}] J_m + I_m \right) \right]. \quad (35)$$

*3) If* $\gamma \neq 1$, $\mu = 1$, *then*

$$\lim_{D \to \infty} \left\langle \log(C_{\tilde{\Phi}_D(\mathbf{x})} + \gamma I_D), \log(C_{\tilde{\Phi}_D(\mathbf{y})} + I_D) \right\rangle_F = \left\langle \log(C_{\Phi(\mathbf{x})} + \gamma I_{\mathcal{H}}), \log(C_{\Phi(\mathbf{y})} + I_{\mathcal{H}}) \right\rangle_{\text{eHS}} + (\log \gamma) \text{tr} \left[ \log \left( \frac{1}{m} J_m K[\mathbf{y}] J_m + I_m \right) \right]. \quad (36)$$

*4) If* $\gamma \neq 1$, $\mu \neq 1$, *then*

$$\lim_{D \to \infty} \left\langle \log(C_{\tilde{\Phi}_D(\mathbf{x})} + \gamma I_D), \log(C_{\tilde{\Phi}_D(\mathbf{y})} + \mu I_D) \right\rangle_F = \pm \infty. \quad (37)$$

*The limiting value on the right hand side of Eq. (37) depends on the sign of* $(\log \gamma)(\log \mu)$. *If* $(\log \gamma)(\log \mu) > 0$, *then this limit is* $+\infty$. *If* $(\log \gamma)(\log \mu) < 0$, *then this limit is* $-\infty$.

**Implications of Theorem 3**. The results stated in Theorem 3 mean that except in the special case $\gamma = \mu = 1$, it is *not* possible to approximate positive definite kernels which are functions of the infinite-dimensional Log-HS inner product by their finite-dimensional counterparts. This is true in particular for linear and polynomial kernels. In fact, for these kernels, in the case $\gamma \neq 1, \mu \neq 1, \gamma > 0, \mu > 0$, the finite-dimensional versions, which are defined in terms of $\left\langle \log(C_{\hat{\Phi}_D(\mathbf{x})} + \gamma I_D), \log(C_{\hat{\Phi}_D(\mathbf{y})} + \mu I_D) \right\rangle_F$, will approach $\pm\infty$ as $D \to \infty$.

In light of Theorems 1, 2, and 3, at the learning step in our numerical work, we employ exclusively shift-invariant kernels ($K_2$ in Fig. 1) which are defined in terms of the approximate Log-HS distance with $\gamma = \mu > 0$.

## Summary of the motivations and theoretical justifications for the proposed model

Having carried out the mathematical analysis for the approximate Log-HS distance and approximate Log-HS inner product between covariance operators, we now summarize the motivations and theoretical justifications for the construction of the proposed model, as described in Algorithm 1 and depicted in Fig. 1.

- For $K_1$, we choose a kernel with an infinite-dimensional feature space $\mathcal{H}$, such as the Gaussian kernel, and corresponding infinite-dimensional covariance operators, in order to capture and exploit the nonlinear correlations between the original features in the input data. In particular, when $K_1$ is shift-invariant, approximate covariance operators can be efficiently computed using random (or quasi-random) Fourier features.

- Given the approximate infinite-dimensional covariance operators corresponding to $K_1$, by Theorem 3, when choosing $K_2$, we exclude kernels defined using the approximate Log-HS inner product, such as linear and polynomial kernels, since the approximate Log-HS inner product generally does not converge to the exact Log-HS inner product and often actually diverges to infinity.

- By Theorems 1 and 2, $K_2$ can be any continuous shift-invariant kernel, such as the Gaussian kernel, defined using the approximate Log-HS distance, on the condition that all covariance operators share the same regularization parameter $\gamma > 0$, since this is the only scenario in which the approximate Log-HS distance converges to a finite value, which is equal to the exact Log-HS distance, as the feature dimension goes to infinity.

- The approximate Log-HS distance is chosen over the approximate affine-invariant distance since (i) it can be used to define positive definite kernels, and (ii) it can be computed much more efficiently than the approximate affine-invariant distance, by the complexity analysis in Section 3.3.

## 5 FEATURE REPRESENTATIONS

In this section, we describe the feature representations used in the experiments. The proposed method has been tested on (i) low-level features using standard state of the art descriptors for image recognition; and (ii) learned features by exploiting the representational power of deep convolutional networks.

In the first representation, for each image, at the pixel at location $(x, y)$, we extract the following low-level features

$$F(x, y) = [x, y, I(x, y), |I_x|, |I_y|, |I_{xx}|, |I_{yy}|,$$
$$R(x, y), G(x, y), B(x, y), |G^{o,s}_{x,y}|], \quad (38)$$

where $I, I_x, I_y, I_{xx}, I_{yy}$ denote the intensity and its first- and second-order derivatives, respectively, $R$, $G$, and $B$ denote the color values, and $G^{o,s}_{x,y}$ denotes the Gabor filter at orientation $o$ and scale $s$. For each dataset in the experiments, the actual features extracted form a subset of Eq. (38) and are the same as those used by the state-of-the-art methods we compare with. If we sample from $m$ pixels, with $n$ features at each pixel, then each image gives rise to a data matrix $\mathbf{x}$ of size $n \times m$, which is used as input to Algorithm 1.

The second representation we tested in the experiments is based on the hypercolumn descriptor [21]. Each image is fed into the pre-trained ConvNet of [26] and its internal representation is used to create the feature representation. Similar to [21], the feature maps *pool2* and *conv4* are concatenated to build the convolutional representation $\mathbf{x}^{conv}$, which is a data matrix of size $640 \times m$ [1], and the map *fc7* is used for the fully-connected representation $\mathbf{x}^{fc}$, which is a 4096-dim vector. Here the convolutional maps *pool2* and *conv4* are resized to a fixed size of $128 \times 128$ pixels and then linearized (our $m$ size, see Section 6).

The convolutional and fully-connected representations are used to build two kernels. For the fully-connected representation $\mathbf{x}^{fc}$, it has been demonstrated empirically in many applications, see e.g. [14], that it suffices to use a linear classifier. Therefore, we used a linear kernel on top of $\mathbf{x}^{fc}$. On the other hand, since the convolutional representation $\mathbf{x}^{conv}$ is extracted from layers that are closer to the pixels, it is expected to yield more variation and noise, thus we expect to extract more information from it by using the covariance operator framework. With the $\mathbf{x}^{conv}$ for each image giving rise to one covariance operator, we can then compute the pairwise approximate Log-HS distances between all the covariance operators. With these distances, we define another positive definite kernel, which is then combined with the linear kernel of the fully-connected representation. In the experiments, we use the point-wise product of the two kernels.

## 6 EXPERIMENTS

In this section, we evaluate the performance of our proposed method, as summarized in Algorithm 1, compared with different state-of-the-art approaches on different challenging datasets for object categorization.

The following methods were evaluated and compared: *LogE*, using the Log-Euclidean metric, which is the method of [22], [23], *E*, using the Euclidean metric, *Stein*, using the Stein (also called Jensen-Bregman LogDet) divergence [11], *HS*, using the Hilbert-Schmidt metric, *LogHS*, using the Log-HS metric [36] induced by the Gaussian kernel ($K_1$ in Fig. 1) , and *Approx LogHS* and *QApprox LogHS* induced by the Gaussian kernel, using the proposed random Fourier and Quasi-random Fourier approximation methods in Section 3, respectively. We set the approximate feature dimension $D = 200$ in Eq. (29) for *Approx LogHS* and *QApprox LogHS* (more details below), for a good trade-off between speed and accuracy. All experiments used LIBSVM [8] for classification, with the Gaussian kernel defined on top of the corresponding metric ($K_2$ in Fig. 1), except with *Stein*. For *Stein*, since the corresponding Gaussian kernel is generally not guaranteed to be positive definite [48], we used the Nearest Neighbor approach as in [11]. We also compared the above methods with CDL

---

1. The dimension $640 \times m$ is given by the following parameters: 640 is given by the concatenation of the two layers pool2 and conv4, with outputs equal to 256 and 384, respectively; $m$ is the pixel-dimension of the convolutional maps.

[53], one of the state-of-the-art approaches in covariance-based learning. The performance for each method is evaluated in terms of classification accuracy unless explicitly specified. All parameters were chosen by cross-validation.

Section 6.1 describes the datasets used in our experiments. We used the same features and experimental protocols of the other state-of-the-art approaches in order to have a fair comparison.

Section 6.2 reports a preliminary analysis which involves the exhaustive testing of all the above methods on a relatively small dataset. One particular aim of this analysis is to demonstrate that the performance of the approximate methods *Approx LogHS* and *QApprox LogHS* compares favorably with that of the exact Log-HS metric. Due to its demanding computational complexity, we did not carry out experiments with the exact Log-HS metric on the larger datasets tested in the subsequent sections.

Section 6.3 is divided into two parts: Section 6.3.1 contains the experiments using low-level features and Section 6.3.2 analyzes the results using the ConvNet representation. In particular, the results were obtained with the ConvNet architecture of [26] pre-trained on ImageNet by the authors of the *Caffe* software [24]. As in [21], the fully-connected representation $\mathbf{x}^{fc}$ is the last fully-connected layer *fc7* and the convolutional representation $\mathbf{x}^{conv}$ is the concatenation of the *pool2* and the *com4* layers. The hypercolumn convolutional maps were re-sized to $128 \times 128$. The dense hypercolumn descriptor was computed on a set of patches of dimension $16 \times 16$ with an overlap of 4 pixels in both $x$ and $y$ directions, for a fair comparison with other descriptors (*e.g.* SIFT [32]). We also tried different patch sizes, ranging from $8 \times 8$ to $32 \times 32$ and observed that the performances were stable with a standard deviation of $0.9\%$. The proposed methods make use of the convolutional representation $\mathbf{x}^{conv}$, the fully-connected representation $\mathbf{x}^{fc}$ and their combination $\mathbf{x}^{conv}+\mathbf{x}^{fc}$. In the experiments, this combination means that the Gaussian kernel obtained from $\mathbf{x}^{conv}$, using the one of the above Log metrics, is multiplied pointwise with the linear kernel applied on $\mathbf{x}^{fc}$. Note that the experiment with $\mathbf{x}^{fc}$ and the Log metrics cannot be performed because $\mathbf{x}^{fc}$ is a vector for each image, and its covariance representation would result in a single value.

## 6.1 Datasets

The **Fish recognition dataset** contains $27,370$ verified fish images acquired from live video by [4]. The dataset contains 23 classes of fish with a variable number of images per class (from 21 to $12,112$), with a mean resolution of $150 \times 120$ pixels. The significant variations in color, pose and illumination inside each class make this dataset very challenging. We conducted two different experiments on this dataset, using the R,G,B features from Eq. (38), as in [36]. The first experiment (named Exp1) is a small scale experiment which used the same protocol and the 10 splits provided by [36], consisting altogether of 345 images, divided in 115 images for training (5 images per class) and 230 images for testing (10 images per class). The second experiment (named Exp2) is carried out on the whole dataset using 10 different random splits, considering, as done previously, 115 images for training (5 images per class), testing on the rest.

The **Virus Classification dataset** [27] contains 15 different virus classes. Each class has 100 images of size $41 \times 41$. For this dataset, following [20], we employed the 25-dimensional feature vector consisting of the intensity, 4 gradients, and 20 Gabor filters in Eq. (38) at 4 orientations and 5 scales. We used the 10 splits provided by the authors in a leave-one-out manner, *i.e.* 9 splits for training and 1 split as query, repeating the procedure 10 times.

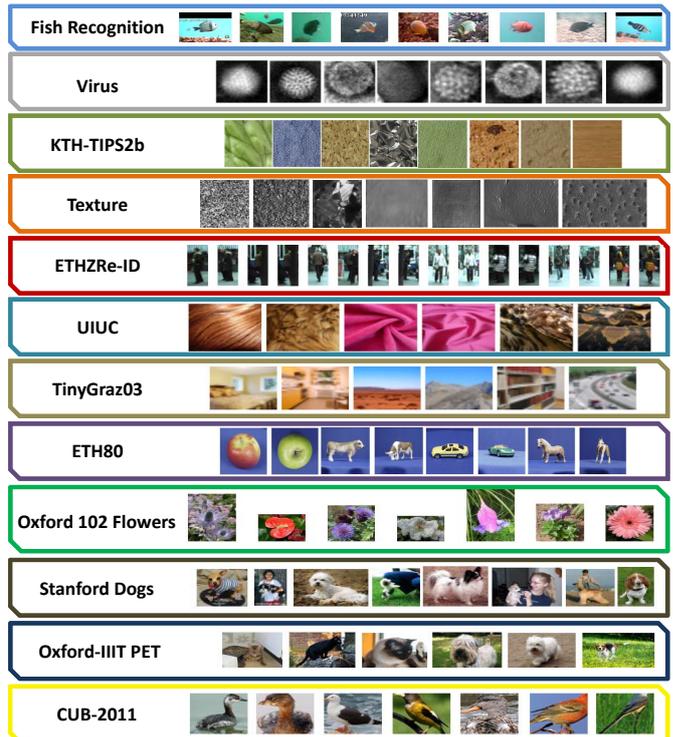

Fig. 2. Sample images for the datasets used in this work. From top to bottom: Fish Recognition dataset [4], Virus Classification dataset [27], KTH-TIPS2b Material dataset [6], Texture dataset [5], [12], ETHZRe-ID dataset [16], UIUC dataset [31], TinyGraz03 dataset [54], ETH80 dataset [29], Oxford 102 Flowers dataset [39], Stanford Dogs dataset [25], Oxford-IIIT PET dataset [40] and Caltech-UCSD Birds 2011 dataset (CUB-2011) [52].

The **KTH-TIPS2b Material dataset** [6] contains images of 11 materials captured under 4 different illuminations, in 3 poses and 9 scales. Therefore, the total number of images is 108 for each sample in a category, with 4 samples per material. For each of the 4 samples, we used 3 images as training and testing on the remaining ones. For this dataset, following [20], we extracted the 23-dimensional feature consisting of the R,G,B values and 20 Gabor filters in Eq. (38) at 4 orientations and 5 scales.

The **Texture dataset** for our experiments was created by combining 111 texture images of the Brodatz dataset [5] and 61 of the CURET dataset [12], as done in [11]. Unfortunately, we were not able to reproduce the experiments in [11] since the exact protocols, i.e. the number of patches extracted from each image and the number of training/testing images, were not specified. We therefore carried out a similar experiment by extracting 150 patches of size $20 \times 20$ from each image, taking 140 as training and 10 as testing, and repeating the entire procedure 10 times. For this dataset, we extracted the 5-dimensional feature vector $[x, y, I, |I_x|, |I_y|]$ as in [11].

For person re-identification, we used two sequences of the **ETHZ dataset** [16], as in [20]. SEQ. #1 contains 83 pedestrians in $4,857$ images. SEQ. #2 contains 35 pedestrians in $1,936$ images. As done in [20], we selected 10 images from each subject for training and used the rest for testing. Following [20], we extracted the 17-dimensional feature vector consisting of $[x, y, R, G, B$ and the first- and second-order derivatives of $R, G, B$, namely $[|\frac{\partial r}{\partial x}|, |\frac{\partial r}{\partial y}|, |\frac{\partial^2 r}{\partial x^2}|, |\frac{\partial^2 r}{\partial y^2}|]$, for $r = R, G, B$. For this dataset the Average Precision metric was used to evaluate the performance.

The **UIUC dataset** [31] is a material categorization dataset that contains 18 different material categories collected *in the wild*. The total number of images for this dataset is 216. The images were collected to have many geometric fine-scale details. Following [17], we extracted the 19-dimensional vector consisting of 3 colors, 4 gradients, and 12 Gabor filters in Eq.(38) at 4 orientations and 3 scales. As in [17], we split the database into training and test sets by randomly assigning half of the images of each class to the training set and testing on the rest. This process was repeated 10 times.

The **TinyGraz03 dataset** [54] contains 1148 indoor and outdoor scenes with an image resolution of $32 \times 32$ pixels. The images are divided in 20 classes with at least 40 samples per class. We used the recommended train/test split provided by the authors. For this dataset, following [17], at each pixel we extracted the 7-dimensional feature vector $[|I_x|, |I_y|, |I_{xx}|, |I_{yy}|, R, G, B]$ from Eq. (38). This dataset is highly challenging and the correct recognition rate achieved by humans is only 30% [54].

The **ETH80 dataset** [29] contains images of eight object categories: apples, cows, cups, dogs, horses, pears, tomatoes, and cars. Each category includes ten object subcategories (eg., various dogs) in 41 orientations, resulting in 410 images per category. We randomly chose 21 images for training, testing on the rest of the data. We repeated this process 10 times. For this dataset, following [22], at each pixel we extracted the 5-dimensional feature vector $[x, y, I(x, y), |I_x|, |I_y|]$ from Eq. (38).

The **Oxford 102 Flowers dataset** [39] contains 8,189 images belonging to 102 flower categories, with each category containing between 40 and 258 images. The split provided is divided into a training set, a validation set and a test set. The training and validation sets each consists of 10 images per class, with the remaining images forming the test set.

The **Stanford Dogs dataset** [25] contains 20,580 images belonging to 120 categories, with 148-252 images per category. The split provided contains a training set of 100 images per category, leaving the rest for testing.

The **Oxford-IIIT PET dataset** [40] contains 7,349 images of cats and dogs of 37 different breeds, with 25 breeds of dogs and 12 breeds of cats. There are approximately 200 images for each breed, split randomly (see [40] for details) into approximately 50 for training, 50 for validation, and 100 for testing.

The **Caltech-UCSD Birds 200-2011 (CUB-200-2011) dataset** [52] contains 11,788 images of 200 bird species. The dataset includes annotations of 15 semantic parts, attributes, and bounding boxes of the objects. In this work, we only used the bounding box of the objects as in [56]. We trained and tested on the splits included with the dataset, which contain around 30 training samples for each species.

## 6.2 Preliminary analysis

First of all, we carried out a comparative experiment on the Fish dataset [4] to analyze the performances of the different Riemannian metrics, namely *LogE*, *LogHS*, *Approx LogHS*, and *QApprox LogHS*, versus the Euclidean and Hilbert-Schmidt metrics, namely *E* and *HS*, and other state-of-the-art approaches. For this experiment, we followed the setup of [36], which uses R,G,B features and a subset of the dataset (Exp1). Next, we performed another comparative experiment using the whole dataset (Exp2), except for *LogHS*, due to its computational complexity, as discussed above.

| Method | Accuracy Exp1 | Accuracy Exp2 |
|---|---|---|
| **Approx LogHS** | **53.91%**(±4.34) | **56.2%**(±2.2) |
| **QApprox LogHS** | **54.30%**(±3.44) | **57.70%**(±1.8) |
| LogHS [36] | 56.74%(±2.87) | N/A |
| HS | 50.17%(±2.17%) | 52.49%(±2.26%) |
| LogE | 42.70%(±3.45) | 46.20%(±1.9) |
| E | 26.87%(±3.52%) | 28.18%(±1.72%) |
| Stein [11] | 43.95%(±4.48) | 40.83%(±7.5) |
| CDL [53] | 41.70%(±3.60) | 42.8%(±2.0) |

TABLE 1
Results on the Fish dataset [4] in terms of classification accuracy. Two different experiments were conducted: Exp1 shows the results on a subset of the dataset as in [36]; Exp2 considers the whole dataset.

For the first experiment, classification accuracies are reported in column two of Table 1, showing the mean and standard deviation values. The first important observation we note is that the Riemannian metrics consistently outperform the Euclidean/Hilbert metrics. In fact, the classification accuracy of *LogHS* is about 6% better compared to that of *HS*. This improvement is clearer when considering the *LogE* and *E* metrics. In this case, in fact, the difference in accuracy between the two metrics is more than 15%.

The second important observation is with regards to the comparison between the two proposed approximations, namely *Approx LogHS* and *QApprox LogHS*, and the exact Log-HS metric, as in [36]. In fact, the first and second rows of Table 1 show that *Approx LogHS* and *QApprox LogHS* give results which are comparable with those using the exact Log-HS metric and substantially better than other methods, namely *LogE*, *Stein*, and *CDL*. At the same time, the approximate methods are much faster to compute than the exact Log-HS metric. To demonstrate this claim, we ran a speed comparison on Exp1 between the two proposed approximate methods and the exact Log-HS metric. Using our MATLAB implementation on an Intel Xeon E5-2650, 2.60 GHz PC, we obtained a speed up of $30\times$ with *QApprox LogHS* (Train: 6.7sec. Test: 18sec.) and more than $50\times$ with *Approx LogHS* (Train: 3.6sec. Test: 9.9sec.) with respect to the baseline Log-HS (Train: 175.7sec. Test: 565.1sec.).

The third column of Table 1 shows the classification accuracies using the whole Fish dataset (Exp2). We see that there is an improvement of 10% and 11.5% for *Approx LogHS* and *QApprox LogHS*, respectively, with respect to *LogE* and an improvement of 15.4% and 16.9% with respect to the *Stein* divergence [11].

Because of the substantial speed up in running time and the relatively small loss in classification accuracy, subsequently we focused solely on the performance of the *Approx LogHS* and *QApprox LogHS* methods for all the other datasets.

## 6.3 Analysis and discussion of results

### 6.3.1 Low-level Features

Table 2 shows the results of *Approx LogHS* and *QApprox LogHS* methods on seven different object classification datasets in comparison with the respective state-of-the-art results. The best result for the Virus dataset is reported in [17] with a classification accuracy of 82.5% (last row). While our classification accuracy is slightly lower than this result, it outperforms all the other competitors (*LogE, Stein and CDL*, by 9.6%, 31.8%, and 12%, respectively). Considering the third column, KTH-TIPS2b Material dataset, our classification accuracy improves the state of the art [20] by 3.5% (third column). The accuracy of the proposed method on the Texture dataset (forth column) is 23.1% higher of the best of the other competitors. The fifth and sixth columns of Table 2 report the results on two different sequences of the ETHZRe-ID dataset. In this case, the obtained improvement is 1.8% over the previous state-of-the-art [20]. Regarding the UIUC and the TinyGraz03 datasets, the improvement over the previous results of [17], are 2.7% and 3%, respectively. Furthermore, our method outperforms the recent state of the art [49] also on the ETH80 dataset with an improvement of 11.4%. All these improvements in classification accuracies demonstrate the effectiveness of our approximation methods. More importantly, we emphasize that our approximate Log-HS formulations are much more computationally efficient than both the RKHS Bregman divergences in [20] and the approximate affine-invariant distance used in [17], as discussed in Section 3.3.

| Method | Virus Acc % | KTH-TIPS2b Acc % | Texture Acc % | ETHZRe-ID | | UIUC Acc % | TinyGraz03 Acc % | ETH80 Acc % |
|---|---|---|---|---|---|---|---|---|
| | | | | Acc-Seq1 % | Acc-Seq2 % | | | |
| **Approx LogHS** | 81.5% (±2.1) | **83.6%** (±5.4) | **76.9%** (±0.5) | **92.0%** (±0.3) | **93.2%** (±0.5) | **50.1%** (±3.7) | **60%** | **95.0%** (±0.5) |
| **QApprox LogHS** | 76.5% (±3.2) | 83.46% (±5.6) | 76.4% (±0.6) | 91.9% (±0.5) | 93.0% (±0.5) | 44.7% (±3.6) | 57% | 94.9% (±0.6) |
| HS | 72.47% (±4.21) | 79.28% (±8.17) | 56.06% (±0.3) | 90.47% (±0.7) | 89.82% (±0.6) | 41.05% (±4.20) | 47% | 93.1% (±0.4) |
| LogE | 71.9% (±4.0) | 74.1% (±7.4) | 52.9% (±0.8) | 89.9% (±0.2) | 91.9% (±0.4) | 37.8% (±2.6) | 40% | 71.1% (±1.0) |
| E | 67.93% (±4.51) | 55.25% (±7.58) | 21.72% (±0.7) | 84.38% (±0.2) | 81.66% (±0.2) | 38.07% (±2.42) | 33% | 64.5% (±0.9) |
| Stein [11] | 49.7% (±4.8) | 73.1% (±8.0) | 38.4% (±0.7) | 89.6% (±0.8) | 90.9% (±0.2) | 27.9% (±1.7) | 24% | 67.5% (±0.4) |
| CDL [53] | 69.5% (±3.1) | 76.3% (±5.11) | 53.8% (±0.5) | 86.8% (±0.6) | 88.8% (±1.2) | 36.3% (±2.0) | 41% | 56.0% (±0.6) |
| SoA | **82.5%** (±2.9) [17] | 80.1% (±4.6) [20] | N/A | 90.2% (±1.0) [20] | 91.4% (±0.8) [20] | 47.4% (±3.1) [17] | 57% [17] | 83.6% (±6.1) [49] |

TABLE 2
Best results obtained on seven different dataset in terms of classification accuracy. The first two rows represent the proposed method using the two-layer kernel machine using *Approx LogHS* and *QApprox LogHS* with Gaussian SVM. The last row represents the state-of-the-art results on each dataset.

Moreover, we carried out a set of experiments to show that the classification accuracy improves when increasing the approximate feature dimension $D$ in Eq. (29), as expected. We decided to test this on the TinyGraz03 dataset, with three different values of $D$: 200, 400, 1000, with the results shown in Table 3.

| Method | D = 200 | D = 400 | D = 1000 |
|---|---|---|---|
| **Approx LogHS** | 57% 3.8s | 59% 20.5s | 60% 240.4s |
| **QApproxLogHS** | 55% 5.6s | 58% 26.7s | 59% 271.7s |

TABLE 3
Results on the TinyGraz03 dataset [54] with increasing values of the approximate feature dimension $D$. We also reported the training time in seconds below each accuracy.

It is worth noting that the differences in terms of classification accuracy between the different $D$ values are not large, however the computational cost grows at a very fast rate, namely $O(D^3)$, as we showed analytically in Section 3.3. Empirically, considering $D = 1000$, the training time for *Approx LogHS* and *QApprox*

*LogHS* is 63 and 48 times slower than $D = 200$, respectively. This suggests that we can exploit the approximation with a relatively low value of $D$ which, while reducing the accuracy by a small amount, results in a large gain in the computational speed.

### 6.3.2 Convolutional Features

In this section, we presents the results obtained with the convolutional features. In Table 4 and Table 5, we show the results of the proposed method on the eight datasets Fish, Oxford 102 Flowers, Standard Dogs, Oxford-IIIT PET, Caltech-UCSD-Birds-2011, UIUC, Virus, and TinyGraz03, in comparison with the respective state-of-the-art results. In the second column, we show a classification accuracy of 75.30% with a standard deviation of ±3.51 of our method on the Fish dataset. In particular, we improved the recognition accuracy of 21.39% with respect to the recent state-of-the-art result [36]. Considering the first row of the Flowers dataset, our method achieves 80.22% classification accuracy which outperforms all previous methods by a minimum of 4% to maximum of 8% without performing any object segmentation [1], [39]. The improvement of our method over the single $\mathbf{x}^{fc}$ baseline is about 8%, demonstrating the complementarity of $\mathbf{x}^{conv}$ and $\mathbf{x}^{fc}$. For the Flower dataset, the highest reported accuracy is 84.6% from [37], in which the authors used a complex pipeline combining object segmentation, multi-scale SIFT descriptors, with spatial pyramid and Fisher vectors. Our approach does not make use of object segmentation and complex feature description, while still obtaining results very close to the ones obtained by their complex pipeline. This further demonstrates the advantages of both our computational framework and the ConvNet feature representation. Regarding the Stanford Dogs dataset, we observe an improvement of the proposed method over the previous state-of-the-art method in [10], from 52.0% to 63.28%, 2% more than the $\mathbf{x}^{fc}$. In the fifth column of Table 4, we present the results obtained on the Oxford-IIIT Pet dataset. The previous state-of-the-art on this dataset was 59.21%, obtained in [40] by performing detection of the head and body of the animal. We instead consider the whole image, without any other information, achieving a recognition accuracy of 75.68%, an improvement of almost 15%. The last column of Table 4 shows the results on the CUB-2011 dataset. In our experiments we used the ground truth position of the bird in order to evaluate the quality of the proposed method when decoupled from the effect of a detector. Considering the state-of-the-art results of [45], [56], we reach a classification accuracy of 60.15%, 2% and 7% more than the previous results (we remark that the result of [56] corresponds to our baseline $\mathbf{x}^{fc}$).

| Method | | Fish Recognition | Oxford 102 Flowers | Stanford Dogs | Oxford-IIIT PET | CUB-2011 |
|---|---|---|---|---|---|---|
| **Approx LogHS** | $\mathbf{x}^{conv}+\mathbf{x}^{fc}$ | **75.30%**(±3.51) | **80.22%** | **63.28%** | **75.68%** | **60.15%** |
| | $\mathbf{x}^{conv}$ | 70.30%(±4.01) | 74.46% | 39.94% | 54.12% | 53.01% |
| | SIFT | 60.30%(±3.42) | 35.69% | 15.94% | 25.66% | 20.81% |
| **QApprox LogHS** | $\mathbf{x}^{conv}+\mathbf{x}^{fc}$ | **74.57%**(±3.59) | 79.87% | 62.7% | **75.20%** | 59.48% |
| | $\mathbf{x}^{conv}$ | 68.48%(±3.91) | 73.51% | 36.76% | 51.31% | 50.45% |
| | SIFT | 58.39%(±3.66) | 35.68% | 15.68% | 25.59% | 20.36% |
| SVM (Baseline) | $\mathbf{x}^{fc}$ | 72.13%(±3.58) | 72.39% | 61.5% | 74.52% | 54.28% |
| State-of-the-art | | 56.74% (±2.87) [36] | 76.7% 74.7% [1], [45] | 52.0% 45.6% [7], [10] | 39.64% 59.21% [40] | 57.94% 53.3% [45], [56] |

TABLE 4
Results obtained using convolutional features. We report the Average Precision accuracy (or standard SVM accuracy when the number of test samples is the same for each class) and the standard deviations (in the case of more than one split, *i.e.* Fish dataset). The proposed method is $\mathbf{x}^{conv}+\mathbf{x}^{fc}$ with *Approx LogHS* and *QApprox LogHS*.

| Method | | UIUC | TinyGraz03 | Virus |
|---|---|---|---|---|
| **Approx LogHS** | $\mathbf{x}^{conv}+\mathbf{x}^{fc}$ | **63.8%**(±3.10) | 71% | **84.13%**(±2.26) |
| | $\mathbf{x}^{conv}$ | 61.12%(±3.63) | 70% | 82.93%(±2.44) |
| **QApprox LogHS** | $\mathbf{x}^{conv}+\mathbf{x}^{fc}$ | **65%**(±2.88) | **73%** | 82.93%(±2.90) |
| | $\mathbf{x}^{conv}$ | 59.08%(±3.87) | 70% | 81.33%(±4.06) |
| SVM (Baseline) | $\mathbf{x}^{fc}$ | 61.02%(±2.94) | 57% | 76%(±2.74) |
| Results from [17] | | 47.4%(±3.1) | 57% | 82.5%(±2.9) |

TABLE 5
Results obtained using convolutional features. We report the Average Precision accuracy (or standard SVM accuracy when the number of test samples is the same for each class) and the standard deviations (in the case of more than one split, *i.e.* UIUC Material and Virus datasets). The proposed method is $\mathbf{x}^{conv}+\mathbf{x}^{fc}$ with *Approx LogHS* and *QApprox LogHS*.

Considering Table 5, in the second column, we show the classification accuracies on the UIUC Material dataset. In this dataset, the highest classification performance obtained by our method is 65% with a standard deviation of ±2.88. In particular, we improved the classification accuracy of 17.6% with respect to the result of [17]. Considering the third row of the TinyGraz dataset, our method achieves 73% classification accuracy, which outperforms [17] by 16%. Finally, the last column presents results on the Virus dataset. In this task, we reached a classification accuracy of 84.13% with a standard deviation of ±2.26, 2% more than the result of [17].

It is interesting to notice that for the TinyGraz03 and the Virus datasets, $\mathbf{x}^{conv}$ gives much better results than $\mathbf{x}^{fc}$. This is due to the fact that the fully-connected features $\mathbf{x}^{fc}$, trained for the task of object classification, do not generalize well to the tasks of scene recognition and cell image classification. This is an expected behavior, because the statistics of these images are very different from the ones which the ConvNet was trained for. On the contrary, $\mathbf{x}^{conv}$ is less prone to this factor, because the convolutional layers are farther from the supervised signal (i.e., the classification output of the ConvNet) and therefore are more general. The proposed method is able to exploit this characteristic of $\mathbf{x}^{conv}$ to our advantage, particularly when the features are combined into $\mathbf{x}^{conv}+\mathbf{x}^{fc}$.

# 7 CONCLUSION, DISCUSSION, AND FUTURE WORK

In this paper, we have presented a novel mathematical and computational framework for visual object recognition using infinite-dimensional covariance operators and their finite-dimensional approximations. Our proposed framework is firmly grounded in the mathematical machinery of Riemannian geometry in the context of kernel methods.

Theoretically, we formulate a finite-dimensional approximation of the Log-HS distance between covariance operators that is substantially faster to compute than the exact Log-HS distance while preserving its discriminative capability. In particular, the approximate Log-HS distance allows us to define approximate shift-invariant kernels, such as the Gaussian kernel, between covariance operators, along with the corresponding kernel methods, that are scalable to large datasets. At the same time, we prove that kernels defined using the infinite-dimensional Log-HS inner product between covariance operators, such as linear and polynomial kernels, can only be approximated by their finite-dimensional counterparts in a very limited scenario. This shows mathematically that the use of shift-invariant kernels between infinite-dimensional covariance operators and their approximations is preferable for large scale applications.

Empirically, we apply our framework to the task of object categorization, using both handcrafted, low-level features and convolutional features. Using numerous challenging, publicly available datasets, we demonstrate that our framework compares very favorably with previous state-of-the-art methods in terms of both classification accuracy and computational complexity.

Given the rigorous mathematical foundation of the current framework and its competitive empirical performance, we believe that this is a promising venue for future research. Some potential directions for future work include the exploration of approximation schemes other than the Fourier feature maps, and the application of our framework to many other problems in computer vision and image processing.

# APPENDIX A
## PROOFS FOR MAIN MATHEMATICAL RESULTS

We need the following preliminary results. Let $\mathcal{H}$ be a separable Hilbert space, with norm $|| \, ||$, and $A : \mathcal{H} \to \mathcal{H}$ be a bounded linear operator. We recall that the operator norm of $A$ is defined to be

$$||A|| = \sup_{x \neq 0} \frac{||Ax||}{||x||}. \tag{39}$$

If $A$ is self-adjoint, compact, and positive, then

$$||A|| = \lambda_{\max}(A), \tag{40}$$

where $\lambda_{\max}(A)$ denotes the largest eigenvalue of $A$. The trace norm of $A$ in this case is given by

$$||A||_{\mathrm{tr}} = \sum_{k=1}^{\infty} \lambda_k(A) = \mathrm{tr}(A). \tag{41}$$

**Lemma 1.** *Let $\mathcal{H}$ be a separable Hilbert space. Let $r \in \mathbb{N}$ be fixed. Let $A \in \mathcal{L}(\mathcal{H})$ be a self-adjoint, positive operator with finite rank $r < \infty$. Then*

$$||A|| \leq ||A||_{\mathrm{HS}} \leq \sqrt{r} ||A||, \tag{42}$$
$$||A|| \leq ||A||_{\mathrm{tr}} \leq r ||A||. \tag{43}$$

*Thus convergences in the $|| \, ||_{\mathrm{HS}}$ norm, the $|| \, ||_{\mathrm{tr}}$ norm, and the $|| \, ||$ norm are all equivalent to each other.*

*Proof.* By definition of the $|| \, ||$ and $|| \, ||_{\mathrm{HS}}$ norms and the finite rank assumption, we have

$$||A||^2 = \lambda_{\max}^2(A) \leq \sum_{j=1}^{r} \lambda_j^2(A) = ||A||_{\mathrm{HS}}^2 \leq r \lambda_{\max}^2(A),$$

from which the first inequality follows. Similarly, for the second inequality, we have

$$||A|| = \lambda_{\max}(A) \leq \sum_{j=1}^{r} \lambda_k(A) = ||A||_{\mathrm{tr}} \leq r \lambda_{\max}(A) = r||A||.$$

This completes the proof of the lemma. $\qquad \square$

**Lemma 2.** *Let $\mathcal{H}$ be a separable Hilbert space. Let $r \in \mathbb{N}$ be fixed. Let $\{A_k\}_{k \in \mathbb{N}}$ be self-adjoint, positive operators of rank at most $r$, such that $\lim_{k \to \infty} ||A_k - A||_{\mathrm{HS}} = 0$. Then*

$$\lim_{k \to \infty} || \log(I + A_k) - \log(I + A) ||_{\mathrm{HS}} = 0. \tag{44}$$

*Proof.* By assumption, the operators in the sequence $(A_k - A)_{k \in \mathbb{N}}$ all have rank at most $2r$. Thus from Lemma 1, the convergence $||A_k - A||_{\mathrm{HS}}$ is equivalent to the convergence $||A_k - A||$. The operators in the sequence $(\log(I + A_k))_{k \in \mathbb{N}}$ are also self-adjoint, positive, and of rank at most $r$. The operators in the sequence $(\log(I + A_k) - \log(I + A))_{k \in \mathbb{N}}$ have rank at most $2r$ and thus the convergence $|| \log(I + A_k) - \log(I + A) ||_{\mathrm{HS}}$ is equivalent to the convergence $|| \log(I + A_k) - \log(I + A) ||$. Thus we have

$$||A_k - A|| \to 0 \Longrightarrow ||A_k - A|| \to 0$$
$$\Longleftrightarrow \lambda_{\max}(A_k) \to \lambda_{\max}(A).$$

It follows that

$$\log(1 + \lambda_{\max}(A_k)) \to \log(1 + \lambda_{\max}(A))$$
$$\Longleftrightarrow || \log(I + A_k) - \log(I + A) || \to 0$$
$$\Longleftrightarrow || \log(I + A_k) - \log(I + A) ||_{\mathrm{HS}} \to 0.$$

This completes the proof of the lemma. $\qquad \square$

**Lemma 3.** *Let $\mathcal{H}$ be a separable Hilbert space. Let $r \in \mathbb{N}$ be fixed. Let $A, \{A_k\}_{k \in \mathbb{N}}$ be self-adjoint, positive operators of rank at most $r$, such that $\lim_{k \to \infty} ||A_k - A||_{\mathrm{HS}} = 0$. Then*

$$\lim_{k \to \infty} \mathrm{tr}[\log(I + A_k) - \log(I + A)] = 0. \tag{45}$$

*Proof.* By assumption, the operators in the sequence $(\log(I + A_k))_{k \in \mathbb{N}}$ are also self-adjoint, positive, and of rank at most $r$. The operators in the sequence $(\log(I + A_k) - \log(I + A))_{k \in \mathbb{N}}$ have rank at most $2r$ and by Lemma 2, $\lim_{k \to \infty} || \log(I + A_k) - \log(I + A) ||_{\mathrm{HS}} = 0$. By Lemma 1, this convergence is equivalent to convergence in the $|| \, ||_{\mathrm{tr}}$ norm. Thus we have

$$|\mathrm{tr}[\log(I + A_k) - \log(I + A)]| \leq || \log(I + A_k) - \log(I + A) ||_{\mathrm{tr}}$$
$$\to 0$$

as $k \to \infty$. This completes the proof of the lemma. $\qquad \square$

**Lemma 4.** *Let $\mathcal{H}$ be a separable Hilbert space. Let $r \in \mathbb{N}$ be fixed. Let $A, \{A_k\}_{k \in \mathbb{N}}$, $B, \{B_k\}_{k \in \mathbb{N}}$ be self-adjoint, positive operators of rank at most $r$, such that $\lim_{k \to \infty} ||A_k - A||_{\mathrm{HS}} = 0$ and $\lim_{k \to \infty} ||B_k - B||_{\mathrm{HS}} = 0$. Then*

$$\lim_{k \to \infty} \mathrm{tr}[\log(I + A_k) \log(I + B_k)] = \mathrm{tr}[\log(I + A) \log(I + B)]. \tag{46}$$

*Proof.* From Lemma 2, we have

$$\lim_{k \to \infty} || \log(I + A_k) - \log(I + A) ||_{\mathrm{HS}} = 0,$$
$$\lim_{k \to \infty} || \log(I + B_k) - \log(I + B) ||_{\mathrm{HS}} = 0.$$

Thus, using Cauchy-Schwarz inequality and the definition $\langle A, B \rangle_{\mathrm{HS}} = \mathrm{tr}(A^* B)$, we obtain

$$|\mathrm{tr}[\log(I + A_k) \log(I + B_k) - \log(I + A) \log(I + B)]|$$
$$= |\mathrm{tr}[\log(I + A_k) \log(I + B_k) - \log(I + A_k) \log(I + B)$$
$$+ \log(I + A_k) \log(I + B) - \log(I + A) \log(I + B)]|$$
$$= |\mathrm{tr}[\log(I + A_k)(\log(I + B_k) - \log(I + B))$$
$$+ (\log(I + A_k) - \log(I + A)) \log(I + B)]|$$
$$= |\langle \log(I + A_k), \log(I + B_k) - \log(I + B) \rangle_{\mathrm{HS}}$$
$$+ \langle \log(I + A_k) - \log(I + A), \log(I + B) \rangle_{\mathrm{HS}}|$$
$$\leq |\langle \log(I + A_k), \log(I + B_k) - \log(I + B) \rangle_{\mathrm{HS}}|$$
$$+ |\langle \log(I + A_k) - \log(I + A), \log(I + B) \rangle_{\mathrm{HS}}|$$
$$\leq || \log(I + A_k) ||_{\mathrm{HS}} || \log(I + B_k) - \log(I + B) ||_{\mathrm{HS}}$$
$$+ || \log(I + A_k) - \log(I + A) ||_{\mathrm{HS}} || \log(I + B) ||_{\mathrm{HS}}.$$

Taking limit on both sides as $k \to \infty$, we obtain

$$\lim_{k \to \infty} \mathrm{tr}[\log(I + A_k) \log(I + B_k) - \log(I + A) \log(I + B)] = 0.$$

This completes the proof of the lemma. $\qquad \square$

**Lemma 5** ( [36]). *Let $\mathcal{H}_1$ and $\mathcal{H}_2$ be two separable Hilbert spaces. Let $A : \mathcal{H}_1 \to \mathcal{H}_2$ and $B : \mathcal{H}_2 \to \mathcal{H}_1$ be two bounded linear operators. Then the nonzero eigenvalues of $BA : \mathcal{H}_1 \to \mathcal{H}_1$ and $AB : \mathcal{H}_2 \to \mathcal{H}_2$, if they exist, are the same.*

In the following, we identify $\mathcal{H}$ with $\ell^2$ and $\mathbb{R}^D$ with a $D$-dimensional subspace of $\ell^2$, that is

$$a = (a_j)_{j=1}^D \in \mathbb{R}^D \Longleftrightarrow a = (a_1, \ldots, a_D, 0, 0, \ldots) \in \ell^2. \tag{47}$$

For the matrices $\mathbf{x} = [x_1, \ldots, x_m]$, $\mathbf{y} = [y_1, \ldots, y_m]$, let $K[\mathbf{x}]$, $K[\mathbf{y}]$, $\tilde{K}_D[\mathbf{x}]$, $\tilde{K}_D[\mathbf{y}]$ be the $m \times m$ Gram matrices defined by

$$(K[\mathbf{x}])_{ij} = K(x_i, x_j), \quad (\tilde{K}_D[\mathbf{x}])_{ij} = \tilde{K}_D(x_i, x_j),$$
$$(K[\mathbf{y}])_{ij} = K(y_i, y_j), \quad (\tilde{K}_D[\mathbf{y}])_{ij} = \tilde{K}_D(y_i, y_j).$$

**Lemma 6.** *Let $\mathcal{H} = \ell^2$, with $\mathbb{R}^D$ identified with a $D$-dimensional subspace of $\mathcal{H}$ as in Eq. (47). Assume that $\lim_{D \to \infty} \tilde{K}_D(x, y) = K(x, y)$ for all pairs $(x, y) \in \mathcal{X} \times \mathcal{X}$. Then*

$$\lim_{D \to \infty} ||C_{\tilde{\Phi}_D(\mathbf{x})} - C_{\Phi(\mathbf{x})}||_{\mathrm{HS}(\mathcal{H})} = 0. \tag{48}$$

*Proof.* Let $A = \frac{1}{\sqrt{m}} \Phi(\mathbf{x}) J_m : \mathbb{R}^m \to \mathcal{H}$, then

$$AA^* = C_{\Phi(\mathbf{x})}, \quad A^* A = \frac{1}{m} J_m K[\mathbf{x}] J_m.$$

By Lemma 5, the nonzero eigenvalues of $C_{\Phi(\mathbf{x})} = AA^*$ are the same as those of $\frac{1}{m} J_m K[\mathbf{x}] J_m = A^* A$. Similarly, the nonzero eigenvalues of $C_{\tilde{\Phi}_D(\mathbf{x})}$ are the same as those of $\frac{1}{m} J_m \tilde{K}_D[\mathbf{x}] J_m$. This also implies that both $C_{\Phi(\mathbf{x})}$ and $C_{\tilde{\Phi}_D(\mathbf{x})}$ have rank at most $m - 1$, since $\text{rank}(J_m) = m - 1$. Since $\lim_{D \to \infty} \tilde{K}_D(x_i, x_j) = K(x_i, x_j) \; \forall (x_i, x_j), \, 1 \le i, j \le m$, we have, as $m \times m$ matrices,

$$\lim_{D \to \infty} ||J_m \tilde{K}_D[\mathbf{x}] J_m - J_m K[\mathbf{x}] J_m||_F = 0.$$

Since $J_m \tilde{K}_D[\mathbf{x}] J_m$ and $J_m K[\mathbf{x}] J_m$ are finite matrices, convergence in the $|| \;||_F$ norm is equivalent to convergence in the operator $|| \;||$ norm. Thus we have

$$\lim_{D \to \infty} ||J_m \tilde{K}_D[\mathbf{x}] J_m - J_m K[\mathbf{x}] J_m|| = 0$$
$$\iff \lim_{D \to \infty} \lambda_{\max}(J_m \tilde{K}_D[\mathbf{x}] J_m) = \lambda_{\max}(J_m K[\mathbf{x}] J_m)$$
$$\iff \lim_{D \to \infty} \lambda_{\max}(C_{\tilde{\Phi}_D(\mathbf{x})}) = \lambda_{\max}(C_{\Phi(\mathbf{x})})$$
$$\iff \lim_{D \to \infty} ||C_{\tilde{\Phi}_D(\mathbf{x})} - C_{\Phi(\mathbf{x})}|| = 0$$
$$\iff \lim_{D \to \infty} ||C_{\tilde{\Phi}_D(\mathbf{x})} - C_{\Phi(\mathbf{x})}||_{\text{HS}(\mathcal{H})} = 0,$$

by Lemma 1, since both $C_{\tilde{\Phi}_D(\mathbf{x})}$, $C_{\Phi(\mathbf{x})}$ have rank at most $m - 1$. This completes the proof of the lemma. $\qquad \square$

***Proof of Theorem 1.*** Consider the expansion

$$\left\| \log\left( C_{\tilde{\Phi}_D(\mathbf{x})} + \gamma I_D \right) - \log\left( C_{\tilde{\Phi}_D(\mathbf{y})} + \mu I_D \right) \right\|_F^2$$
$$= \left\| \log\left( \frac{C_{\tilde{\Phi}_D(\mathbf{x})}}{\gamma} + I_D \right) - \log\left( \frac{C_{\tilde{\Phi}_D(\mathbf{y})}}{\mu} + I_D \right) + (\log \frac{\gamma}{\mu}) I_D \right\|_F^2$$
$$= \left\| \log\left( \frac{C_{\tilde{\Phi}_D(\mathbf{x})}}{\gamma} + I_D \right) - \log\left( \frac{C_{\tilde{\Phi}_D(\mathbf{y})}}{\mu} + I_D \right) \right\|_F^2$$
$$+ 2(\log \frac{\gamma}{\mu}) \text{tr}\left( \log\left( \frac{C_{\tilde{\Phi}_D(\mathbf{x})}}{\gamma} + I_D \right) - \log\left( \frac{C_{\tilde{\Phi}_D(\mathbf{y})}}{\mu} + I_D \right) \right)$$
$$+ (\log \frac{\gamma}{\mu})^2 D. \tag{49}$$

With $\mathbb{R}^D$ identified as a subspace of $\mathcal{H} = \ell^2$, we have by Lemma 6 (with the scaling factors $\gamma, \mu$), that

$$\lim_{D \to \infty} \left\| \frac{C_{\tilde{\Phi}_D(\mathbf{x})}}{\gamma} - \frac{C_{\Phi(\mathbf{x})}}{\gamma} \right\|_{\text{HS}}^2 = 0, \; \lim_{D \to \infty} \left\| \frac{C_{\tilde{\Phi}_D(\mathbf{y})}}{\mu} - \frac{C_{\Phi(\mathbf{y})}}{\mu} \right\|_{\text{HS}}^2 = 0.$$

By Lemma 3, we have

$$\lim_{D \to \infty} \text{tr}\left( \log\left( \frac{C_{\tilde{\Phi}_D(\mathbf{x})}}{\gamma} + I_D \right) \right) = \text{tr}\left( \log\left( \frac{C_{\Phi(\mathbf{x})}}{\gamma} + I_{\mathcal{H}} \right) \right)$$
$$= \text{tr}\left[ \log\left( \frac{1}{\gamma m} J_m K[\mathbf{x}] J_m + I_m \right) \right],$$
$$\lim_{D \to \infty} \text{tr}\left( \log\left( \frac{C_{\tilde{\Phi}_D(\mathbf{y})}}{\mu} + I_D \right) \right) = \text{tr}\left( \log\left( \frac{C_{\Phi(\mathbf{y})}}{\mu} + I_{\mathcal{H}} \right) \right)$$
$$= \text{tr}\left[ \log\left( \frac{1}{\mu m} J_m K[\mathbf{y}] J_m + I_m \right) \right].$$

Since these two quantities are both finite, for $\gamma \ne \mu$, as $D \to \infty$, clearly the right hand side of Eq. (49) goes to infinity. This gives us the desired limit. $\qquad \square$

***Proof of Theorem 2.*** Without loss of generality, we identify $\mathcal{H}$ with $\ell^2$ as above and identify $\mathbb{R}^D$ with a $D$-dimensional subspace of $\ell^2$ as in Eq. (47). When $\gamma = \mu$, we have

$$|| \log(C_{\Phi(\mathbf{x})} + \gamma I_{\mathcal{H}}) - \log(C_{\Phi(\mathbf{y})} + \gamma I_{\mathcal{H}}) ||_{\text{eHS}}^2$$
$$= \left\| \log\left( \frac{C_{\Phi(\mathbf{x})}}{\gamma} + I_{\mathcal{H}} \right) - \log\left( \frac{C_{\Phi(\mathbf{y})}}{\gamma} + I_{\mathcal{H}} \right) \right\|_{\text{HS}}^2$$
$$= \left\| \log\left( \frac{C_{\Phi(\mathbf{x})}}{\gamma} + I_{\mathcal{H}} \right) \right\|_{\text{HS}}^2 + \left\| \log\left( \frac{C_{\Phi(\mathbf{y})}}{\gamma} + I_{\mathcal{H}} \right) \right\|_{\text{HS}}^2$$
$$- 2\text{tr}\left[ \log\left( \frac{C_{\Phi(\mathbf{x})}}{\gamma} + I_{\mathcal{H}} \right) \log\left( \frac{C_{\Phi(\mathbf{y})}}{\gamma} + I_{\mathcal{H}} \right) \right].$$

It follows from Lemma 5 that the first term is

$$\left\| \log\left( \frac{1}{\gamma} C_{\Phi(\mathbf{x})} + I_{\mathcal{H}} \right) \right\|_{\text{HS}}^2 = \left\| \log\left( \frac{1}{\gamma m} \Phi(\mathbf{x}) J_m^2 \Phi(\mathbf{x})^* + I_{\mathcal{H}} \right) \right\|_{\text{HS}}^2$$
$$= \left\| \log\left( \frac{1}{\gamma m} J_m \Phi(\mathbf{x})^* \Phi(\mathbf{x}) J_m + I_m \right) \right\|_{\text{HS}}^2$$
$$= \left\| \log\left( \frac{1}{\gamma m} J_m K[\mathbf{x}] J_m + I_m \right) \right\|_{\text{HS}}^2$$
$$= \text{tr}\left[ \log\left( \frac{1}{\gamma m} J_m K[\mathbf{x}] J_m + I_m \right) \right]^2.$$

Similarly, the second term is

$$\left\| \log\left( \frac{1}{\gamma} C_{\Phi(\mathbf{y})} + I_{\mathcal{H}} \right) \right\|_{\text{HS}}^2 = \text{tr}\left[ \log\left( \frac{1}{\gamma m} J_m K[\mathbf{y}] J_m + I_m \right) \right]^2.$$

Thus we have

$$|| \log(C_{\Phi(\mathbf{x})} + \gamma I_{\mathcal{H}}) - \log(C_{\Phi(\mathbf{y})} + \gamma I_{\mathcal{H}}) ||_{\text{eHS}}^2$$
$$= \text{tr}\left[ \log\left( \frac{1}{\gamma m} J_m K[\mathbf{x}] J_m + I_m \right) \right]^2$$
$$+ \text{tr}\left[ \log\left( \frac{1}{\gamma m} J_m K[\mathbf{y}] J_m + I_m \right) \right]^2$$
$$- 2\text{tr}\left[ \log\left( \frac{C_{\Phi(\mathbf{x})}}{\gamma} + I_{\mathcal{H}} \right) \log\left( \frac{C_{\Phi(\mathbf{y})}}{\gamma} + I_{\mathcal{H}} \right) \right]. \tag{50}$$

Similarly,

$$|| \log(C_{\tilde{\Phi}_D(\mathbf{x})} + \gamma I_D) - \log(C_{\tilde{\Phi}_D(\mathbf{y})} + \gamma I_D) ||_F^2 \tag{51}$$
$$= \text{tr}\left[ \log\left( \frac{1}{\gamma m} J_m \tilde{K}_D[\mathbf{x}] J_m + I_m \right) \right]^2$$
$$+ \text{tr}\left[ \log\left( \frac{1}{\gamma m} J_m \tilde{K}_D[\mathbf{y}] J_m + I_m \right) \right]^2$$
$$- 2\text{tr}\left[ \log\left( \frac{C_{\tilde{\Phi}_D(\mathbf{x})}}{\gamma} + I_D \right) \log\left( \frac{C_{\tilde{\Phi}_D(\mathbf{y})}}{\gamma} + I_D \right) \right]. \tag{52}$$

With $\mathbb{R}^D$ identified as a subspace of $\mathcal{H} = \ell^2$, we have by Lemma 6 (with the scaling factor $\gamma$), that

$$\lim_{D \to \infty} \left\| \frac{C_{\tilde{\Phi}_D(\mathbf{x})}}{\gamma} - \frac{C_{\Phi(\mathbf{x})}}{\gamma} \right\|_{\text{HS}}^2 = 0, \; \lim_{D \to \infty} \left\| \frac{C_{\tilde{\Phi}_D(\mathbf{y})}}{\gamma} - \frac{C_{\Phi(\mathbf{y})}}{\gamma} \right\|_{\text{HS}}^2 = 0,$$

with the operators $C_{\tilde{\Phi}_D(\mathbf{x})}$, $C_{\Phi(\mathbf{x})}$, $C_{\tilde{\Phi}_D(\mathbf{y})}$, $C_{\Phi(\mathbf{y})}$ all having rank at most $m - 1$. It thus follows from Lemma 4 that

$$\lim_{D \to \infty} \text{tr}\left[ \log\left( \frac{C_{\tilde{\Phi}_D(\mathbf{x})}}{\gamma} + I_D \right) \log\left( \frac{C_{\tilde{\Phi}_D(\mathbf{y})}}{\gamma} + I_D \right) \right]$$
$$= \text{tr}\left[ \log\left( \frac{C_{\Phi(\mathbf{x})}}{\gamma} + I_{\mathcal{H}} \right) \log\left( \frac{C_{\Phi(\mathbf{y})}}{\gamma} + I_{\mathcal{H}} \right) \right]. \tag{53}$$

Similarly, since $\lim_{D \to \infty} \tilde{K}_D(x_i, x_j) = K(x_i, x_j)$ for all pairs $(x_i, x_j)$ and $\lim_{D \to \infty} \tilde{K}_D(y_i, y_j) = K(y_i, y_j)$ for all pairs $(y_i, y_j)$, $1 \le i, j \le m$, we have, as $m \times m$ matrices,

$$\lim_{D \to \infty} ||J_m \tilde{K}_D[\mathbf{x}] J_m - J_m K[\mathbf{x}] J_m||_F = 0,$$
$$\lim_{D \to \infty} ||J_m \tilde{K}_D[\mathbf{y}] J_m - J_m K[\mathbf{y}] J_m||_F = 0.$$

It also follows from Lemma 4 that

$$\lim_{D \to \infty} \operatorname{tr} \left[ \log \left( \frac{1}{\gamma m} J_m \hat{K}_D[\mathbf{x}] J_m + I_m \right) \right]^2$$
$$= \operatorname{tr} \left[ \log \left( \frac{1}{\gamma m} J_m K[\mathbf{x}] J_m + I_m \right) \right]^2, \tag{54}$$

$$\lim_{D \to \infty} \operatorname{tr} \left[ \log \left( \frac{1}{\gamma m} J_m \hat{K}_D[\mathbf{y}] J_m + I_m \right) \right]^2$$
$$= \operatorname{tr} \left[ \log \left( \frac{1}{\gamma m} J_m K[\mathbf{y}] J_m + I_m \right) \right]^2. \tag{55}$$

Combining Eqs. (50), (51), (53), (54), and (55), we obtain

$$\lim_{D \to \infty} || \log(C_{\hat{\Phi}_D(\mathbf{x})} + \gamma I_D) - \log(C_{\hat{\Phi}_D(\mathbf{y})} + \gamma I_D) ||_F^2$$
$$= || \log(C_{\Phi(\mathbf{x})} + \gamma I_{\mathcal{H}}) - \log(C_{\Phi(\mathbf{y})} + \gamma I_{\mathcal{H}}) ||_{\text{eHS}}^2.$$

This completes the proof of the Theorem. □

***Proof of Theorem 3.*** By definition of the Log-HS inner product, for $\dim(\mathcal{H}) = \infty$, we have

$$\langle (C_{\Phi(\mathbf{x})} + \gamma I_{\mathcal{H}}), (C_{\Phi(\mathbf{y})} + \mu I_{\mathcal{H}}) \rangle_{\text{logHS}} \tag{56}$$
$$= \langle \log (C_{\Phi(\mathbf{x})} + \gamma I_{\mathcal{H}}), \log (C_{\Phi(\mathbf{y})} + \mu I_{\mathcal{H}}) \rangle_{\text{eHS}}$$
$$= \left\langle \log \left( \frac{C_{\Phi(\mathbf{x})}}{\gamma} + I \right) + (\log \gamma) I, \log \left( \frac{C_{\Phi(\mathbf{y})}}{\mu} + I \right) + (\log \mu) I \right\rangle_{\text{eHS}}$$
$$= \left\langle \log \left( \frac{C_{\Phi(\mathbf{x})}}{\gamma} + I_{\mathcal{H}} \right), \log \left( \frac{C_{\Phi(\mathbf{y})}}{\mu} + I_{\mathcal{H}} \right) \right\rangle_{\text{HS}} + (\log \gamma)(\log \mu)$$
$$= \operatorname{tr} \left[ \log \left( \frac{C_{\Phi(\mathbf{x})}}{\gamma} + I_{\mathcal{H}} \right) \log \left( \frac{C_{\Phi(\mathbf{y})}}{\mu} + I_{\mathcal{H}} \right) \right] + (\log \gamma)(\log \mu).$$

On the other hand, for the approximate covariance operators $C_{\hat{\Phi}_D(\mathbf{x})}, C_{\hat{\Phi}_D(\mathbf{y})}$, we have

$$\left\langle \log(C_{\hat{\Phi}_D(\mathbf{x})} + \gamma I_D), \log(C_{\hat{\Phi}_D(\mathbf{y})} + \mu I_D) \right\rangle_F$$
$$= \left\langle \log \left( \frac{C_{\hat{\Phi}_D(\mathbf{x})}}{\gamma} + I_D \right) + (\log \gamma) I_D \right.,$$
$$\left. \log \left( \frac{C_{\hat{\Phi}_D(\mathbf{y})}}{\mu} + I_D \right) + (\log \mu) I_D \right\rangle_F$$
$$= \operatorname{tr} \left[ \log \left( \frac{C_{\hat{\Phi}_D(\mathbf{x})}}{\gamma} + I_D \right) \log \left( \frac{C_{\hat{\Phi}_D(\mathbf{y})}}{\mu} + I_D \right) \right]$$
$$+ (\log \gamma) \operatorname{tr} \left[ \log \left( \frac{C_{\hat{\Phi}_D(\mathbf{y})}}{\mu} + I_D \right) \right]$$
$$+ (\log \mu) \operatorname{tr} \left[ \log \left( \frac{C_{\hat{\Phi}_D(\mathbf{x})}}{\gamma} + I_D \right) \right] + (\log \gamma)(\log \mu) D. \tag{57}$$

Similar to the proof of Theorem 2, with $\mathbb{R}^D$ identified as a subspace of $\mathcal{H} = \ell^2$, we have by Lemma 6 (with the scaling factors $\gamma$ and $\mu$, respectively), that

$$\lim_{D \to \infty} \left\| \frac{C_{\hat{\Phi}_D(\mathbf{x})}}{\gamma} - \frac{C_{\Phi(\mathbf{x})}}{\gamma} \right\|_{\text{HS}} = 0, \lim_{D \to \infty} \left\| \frac{C_{\hat{\Phi}_D(\mathbf{y})}}{\mu} - \frac{C_{\Phi(\mathbf{y})}}{\mu} \right\|_{\text{HS}} = 0,$$

with the operators $C_{\hat{\Phi}_D(\mathbf{x})}, C_{\Phi(\mathbf{x})}, C_{\hat{\Phi}_D(\mathbf{y})}, C_{\Phi(\mathbf{y})}$ all having rank at most $m - 1$. It thus follows from Lemma 4 that

$$\lim_{D \to \infty} \operatorname{tr} \left[ \log \left( \frac{C_{\hat{\Phi}_D(\mathbf{x})}}{\gamma} + I_D \right) \log \left( \frac{C_{\hat{\Phi}_D(\mathbf{y})}}{\mu} + I_D \right) \right]$$
$$= \operatorname{tr} \left[ \log \left( \frac{C_{\Phi(\mathbf{x})}}{\gamma} + I_{\mathcal{H}} \right) \log \left( \frac{C_{\Phi(\mathbf{y})}}{\mu} + I_{\mathcal{H}} \right) \right]. \tag{58}$$

Consider first the case $\gamma = \mu = 1$, so that $\log(\gamma) = \log(\mu) = 0$, then combining (56), (57), and (58), we obtain

$$\lim_{D \to \infty} \left\langle \log(C_{\hat{\Phi}_D(\mathbf{x})} + I_D), \log(C_{\hat{\Phi}_D(\mathbf{y})} + I_D) \right\rangle_F$$
$$= \left\langle \log \left( C_{\Phi(\mathbf{x})} + I_{\mathcal{H}} \right), \log \left( C_{\Phi(\mathbf{y})} + I_{\mathcal{H}} \right) \right\rangle_{\text{eHS}}.$$

Consider next the case $\gamma = 1$, $\mu \neq 1$. As in the proof of Theorem 1, we have

$$\lim_{D \to \infty} \operatorname{tr} \left( \log \left( \frac{C_{\hat{\Phi}_D(\mathbf{x})}}{\gamma} + I_D \right) \right) = \operatorname{tr} \left( \log \left( \frac{C_{\Phi(\mathbf{x})}}{\gamma} + I_{\mathcal{H}} \right) \right)$$
$$= \operatorname{tr} \left[ \log \left( \frac{1}{\gamma m} J_m K[\mathbf{x}] J_m + I_m \right) \right].$$

Thus combining (56), (57), and (58), we obtain

$$\lim_{D \to \infty} \left\langle \log(C_{\hat{\Phi}_D(\mathbf{x})} + I_D), \log(C_{\hat{\Phi}_D(\mathbf{y})} + \mu I_D) \right\rangle_F$$
$$= \left\langle \log \left( C_{\Phi(\mathbf{x})} + I_{\mathcal{H}} \right), \log \left( C_{\Phi(\mathbf{y})} + \mu I_{\mathcal{H}} \right) \right\rangle_{\text{eHS}}$$
$$+ (\log \mu) \operatorname{tr} \left[ \log \left( \frac{1}{m} J_m K[\mathbf{x}] J_m + I_m \right) \right].$$

The case $\gamma \neq 1$, $\mu = 1$ is proved similarly. Finally, for the case $\gamma \neq 1$, $\mu \neq 1$, we see that on the right hand side of (57), the first three terms all approach finite limits as $D \to \infty$. However, the last term, namely $(\log \gamma)(\log \mu) D$, clearly approaches either $+\infty$ or $-\infty$ as $D \to \infty$, depending on the sign of $(\log \gamma)(\log \mu)$. This completes the proof. □

# APPENDIX B
# THE HILBERT-SCHMIDT DISTANCE BETWEEN CO-VARIANCE OPERATORS

For completeness, in this section we provide the mathematical expression for the Hilbert-Schmidt distance between two covariance operators. This was used for carrying out the corresponding experiments which compare the empirical performance of the Log-HS distance and its approximations with the Hilbert-Schmidt distance. Let $K$ be a positive definite kernel on an arbitrary non-empty set $\mathcal{X}$ and $\mathcal{H}$ be a corresponding Hilbert feature space, which can be taken to be its RKHS $\mathcal{H}_K$. Let $C_{\Phi(\mathbf{x})}$ and $C_{\Phi(\mathbf{y})}$ be the covariance operators on $\mathcal{H}$ corresponding to two $n \times m$ data matrices $\mathbf{x}$ and $\mathbf{y}$, respectively, sampled from $\mathcal{X}$. Following [36], let $K[\mathbf{x}]$, $K[\mathbf{y}]$, and $K[\mathbf{x}, \mathbf{y}]$ denote the $m \times m$ Gram matrices defined by

$$(K[\mathbf{x}])_{ij} = K(x_i, x_j), \ (K[\mathbf{y}])_{ij} = K(y_i, y_j),$$
$$(K[\mathbf{x}, \mathbf{y}])_{ij} = K(x_i, y_j), \ 1 \leq i, j \leq m.$$

The Gram matrices and the covariance operators are related by

$$\Phi(\mathbf{x})^* \Phi(\mathbf{x}) = K[\mathbf{x}], \quad \Phi(\mathbf{y})^* \Phi(\mathbf{y}) = K[\mathbf{y}],$$
$$\Phi(\mathbf{x})^* \Phi(\mathbf{y}) = K[\mathbf{x}, \mathbf{y}].$$

Here $\Phi(\mathbf{x})^*$ denotes the adjoint operator of $\Phi(\mathbf{x})$ in the case $\dim(\mathcal{H}_K) = \infty$ and the transpose of $\Phi(\mathbf{x})$ when $\dim(\mathcal{H}_K) < \infty$.

**Lemma 7.** *The Hilbert-Schmidt distances between the two covariance operators $C_{\Phi(\mathbf{x})}$ and $C_{\Phi(\mathbf{y})}$ is given by*

$$||C_{\Phi(\mathbf{x})} - C_{\Phi(\mathbf{y})}||_{\text{HS}}^2 = \frac{1}{m^2} \langle J_m K[\mathbf{x}], K[\mathbf{x}] J_m \rangle_F$$
$$- \frac{2}{m^2} \langle J_m K[\mathbf{x}, \mathbf{y}], K[\mathbf{x}, \mathbf{y}] J_m \rangle_F$$
$$+ \frac{1}{m^2} \langle J_m K[\mathbf{y}], K[\mathbf{y}] J_m \rangle_F. \tag{59}$$

***Proof of Lemma 7.*** By definition of the Hilbert-Schmidt norm and property of the trace operation, we have

$$||C_{\Phi(\mathbf{x})} - C_{\Phi(\mathbf{y})}||_{\mathrm{HS}}^2 = \left\| \frac{1}{m}\Phi(\mathbf{x})J_m\Phi(\mathbf{x})^* - \frac{1}{m}\Phi(\mathbf{y})J_m\Phi(\mathbf{y})^* \right\|_{\mathrm{HS}}^2$$

$$= \frac{1}{m^2}||\Phi(\mathbf{x})J_m\Phi(\mathbf{x})^*||_{\mathrm{HS}}^2 - \frac{2}{m^2}\langle\Phi(\mathbf{x})J_m\Phi(\mathbf{x})^*, \Phi(\mathbf{y})J_m\Phi(\mathbf{y})^*\rangle_{\mathrm{HS}}$$
$$+ \frac{1}{m^2}||\Phi(\mathbf{y})J_m\Phi(\mathbf{y})^*||_{\mathrm{HS}}^2$$

$$= \frac{1}{m^2}\mathrm{tr}[\Phi(\mathbf{x})J_m\Phi(\mathbf{x})^*\Phi(\mathbf{x})J_m\Phi(\mathbf{x})^*]$$
$$- \frac{2}{m^2}\mathrm{tr}[\Phi(\mathbf{x})J_m\Phi(\mathbf{x})^*\Phi(\mathbf{y})J_m\Phi(\mathbf{y})^*]$$
$$+ \frac{1}{m^2}\mathrm{tr}[\Phi(\mathbf{y})J_m\Phi(\mathbf{y})^*\Phi(\mathbf{y})J_m\Phi(\mathbf{y})^*]$$

$$= \frac{1}{m^2}\mathrm{tr}[(K[\mathbf{x}]J_m)^2 - 2K[\mathbf{y}, \mathbf{x}]J_mK[\mathbf{x}, \mathbf{y}]J_m + (K[\mathbf{y}]J_m)^2]$$

$$= \frac{1}{m^2}[\langle J_mK[\mathbf{x}], K[\mathbf{x}]J_m\rangle_F - 2\langle J_mK[\mathbf{x}, \mathbf{y}], K[\mathbf{x}, \mathbf{y}]J_m\rangle_F$$
$$+ \langle J_mK[\mathbf{y}], K[\mathbf{y}]J_m\rangle_F].$$

This completes the proof of the lemma. □

# APPENDIX C
# THE QUASI-RANDOM FOURIER FEATURES

In this section, we describe in more detail the Quasi-random Fourier features approach proposed recently by [55]. Consider the expression of the kernel $K(x, y) = k(x − y)$ by Bochner's theorem in Eq. (24), namely

$$k(x - y) = \int_{\mathbb{R}^n} e^{-i\langle\omega, x-y\rangle} d\rho(\omega)$$
$$= \int_{\mathbb{R}^n} \rho(\omega)\phi_\omega(x)\overline{\phi_\omega(y)}d\omega, \text{ where } \phi_\omega(x) = e^{-i\langle\omega, x\rangle}.$$

The Random Fourier feature maps arise from the Monte-Carlo approximation of the integral in Eq. (24), using a *random* set of points $\omega_j$'s sampled according to the distribution $\rho$. The Quasi-random Fourier features approach is based instead on the methodology of Quasi-Monte Carlo integration [13], in which the $\omega_j$'s are *deterministic* points arising from a *low-discrepancy* sequence in $[0, 1]^n$ (see below for more details).

Assume that the distribution $\rho$ in Eq. (24) has the product form $\rho(\omega) = \prod_{j=1}^n \rho_j(\omega_j)$. Assume that each component cumulative distribution function $\psi_j(x_j) = \int_{-\infty}^{x_j} \rho_j(z_j)dz_j$ is strictly increasing, so that the inverse functions $\psi_j^{-1} : [0, 1] \rightarrow \mathbb{R}$ are all well-defined. Let $\psi : \mathbb{R}^n \rightarrow [0, 1]^n$ be defined by $\psi(x) = \psi(x_1, \ldots, x_n) = (\psi_1(x_1), \ldots, \psi_n(x_n))$. Then its inverse function $\psi^{-1} : [0, 1]^n \rightarrow \mathbb{R}^n$ is well-defined and is given by $\psi^{-1}(z) = \psi^{-1}(z_1, \ldots, z_n) = (\psi_1^{-1}(z_1), \ldots, \psi_n^{-1}(z_n))$. With the change of variable $\omega = \psi^{-1}(t)$, the integral in Eq. (24) becomes

$$\int_{\mathbb{R}^n} e^{-i\langle\omega, x-y\rangle} \rho(\omega)d\omega = \int_{[0,1]^n} e^{-i\langle\psi^{-1}(t), x-y\rangle} dt. \quad (60)$$

Instead of approximating the left hand side of Eq. (60) using a random set of points $\{\omega_j\}_{j=1}^D$ in $\mathbb{R}^n$ sampled according to $\rho$, in the Quasi-Monte Carlo approach, one approximates the right hand side using a deterministic, low-discrepancy sequence of points $\{t_j\}_{j=1}^D$ in $[0, 1]^n$. With the corresponding deterministic sequence

$$\omega_j = \psi^{-1}(t_j), \quad 1 \leq j \leq D, \quad (61)$$

we construct the Fourier feature map as described by Eqs. (27), (28), and (29), just as in the case of random Fourier features. In our experiments, $\{t_j\}_{j=1}^D$ is a Halton sequence, whose implementation is readily available in MATLAB[2].

**Low-discrepancy sequences.** We now briefly review the concept of *low-discrepancy sequences* in Quasi-Monte Carlo methods. For a comprehensive treatment, we refer to [38]. Let $n \in \mathbb{N}$ be fixed. Let $I^n = [0, 1)^n$ and denote its closure by $\overline{I}^n = [0, 1]^n$. For an integrable function $f$ in $\overline{I}^n$, we consider the approximation

$$\int_{\overline{I}^n} f(u)du \approx \frac{1}{N}\sum_{j=1}^N f(x_j) \quad (62)$$

using a deterministic set of points $P = (x_1, \ldots, x_N)$, which are part of an infinite sequence $(x_j)_{j\in\mathbb{N}}$ in $\overline{I}^n$, such that

$$\lim_{N\to\infty} \left| \frac{1}{N}\sum_{j=1}^N f(x_j) - \int_{\overline{I}^n} f(u)du \right| = 0. \quad (63)$$

This convergence of the integration error can be measured via the concept of *discrepancy* as follows. Let $N$ be fixed. For an arbitrary set $B \subset \overline{I}^n$, define the counting function

$$A(B; P) = \sum_{j=1}^N \chi_B(x_j), \quad (64)$$

where $\chi_B$ denotes the characteristic function for $B$. Thus $A(B; P)$ denotes the number of points in $P$ that lie in the set $B$. Let $\mathcal{B}$ be a non-empty family of Lebesgue-measurable subsets of $\overline{I}^n$. The *discrepancy* of $P$ with respect to $\mathcal{B}$ is defined by

$$D_N(\mathcal{B}; P) = \sup_{B\in\mathcal{B}} \left| \frac{A(B; P)}{N} - \mathrm{vol}(B) \right|, \quad (65)$$

with $\mathrm{vol}(B)$ denoting the volume of $B$ with respect to the Lebesgue measure. The *star discrepancy* $D_N^*(P)$ is defined by

$$D_N^*(P) = D_N(\mathcal{J}^*; P), \quad (66)$$

where $\mathcal{J}^*$ denotes the family of all subintervals of $I^n$ of the form $\prod_{j=1}^n [0, x_j)$. The star discrepancy and the integration error are related via the Koksma- Hlawka inequality, as follows. Define the *variation of $f$ on $\overline{I}^n$ in the sense of Hardy-Krause* by

$$V(f) = \sum_{k=1}^n \sum_{1\leq i_1\leq \cdots \leq i_k\leq n} \int_{[0,1]^k} \left| \frac{\partial^k f}{\partial u_{i_1}\cdots\partial u_{i_k}} \right| du_{i_1}\ldots du_{i_k}. \quad (67)$$

**Theorem 4 (Koksma-Hlawka inequality).** *If $f$ has bounded variation $V(f)$ on $\overline{I}^n$ in the sense of Hardy-Krause, then for any set $(x_1, \ldots, x_N)$ in $I^n$,*

$$\left| \frac{1}{N}\sum_{j=1}^N f(x_j) - \int_{\overline{I}^n} f(u)du \right| \leq V(f)D_N^*(x_1, \ldots, x_N). \quad (68)$$

By Theorem 4, to achieve a small integration error, we need a sequence $(x_j)_{j\in\mathbb{N}}$ with *low discrepancy* $D_N^*(x_1, \ldots, x_N) \to 0$ as $N \to \infty$. Some examples of low-discrepancy sequences are Halton and Sobol' sequences (we refer to [13], [38] for the detailed constructions of these and other sequences). The Halton sequence in particular satisfies $D_N^*(x_1, \ldots, x_N) = C(n)\frac{(\log N)^n}{N}$ for $N \geq 2$.

**The Gaussian case.** We now give the explicit expressions for the functions $\psi$ and $\psi^{-1}$ defined above in the case of the Gaussian kernel. It suffices to consider the one-dimensional case, since the multivariate case is defined componentwise. For $K(x, y) = e^{-\frac{(x-y)^2}{\sigma^2}}$, we have $\rho(z) = \frac{\sigma}{2\sqrt{\pi}}e^{-\frac{\sigma^2 z^2}{4}}$. Recall the Gaussian error function $\mathrm{erf}(x) = \frac{2}{\sqrt{\pi}}\int_0^x e^{-z^2}dz$ and the complementary Gaussian error function $\mathrm{erfc}(x) = \frac{2}{\sqrt{\pi}}\int_x^\infty e^{-z^2}dz = 1 - \mathrm{erf}(x)$. By definition, the cumulative distribution function $\psi$ for $\rho$ is

$$\psi(x) = \int_{-\infty}^x \rho(z)dz = 1 - \int_x^\infty \rho(z)dz = 1 - \frac{\sigma}{2\sqrt{\pi}}\int_x^\infty e^{-\frac{\sigma^2 z^2}{4}}dz$$
$$= 1 - \frac{1}{\sqrt{\pi}}\int_{\frac{x\sigma}{2}}^\infty e^{-u^2}du = 1 - \frac{1}{2}\mathrm{erfc}\left(\frac{x\sigma}{2}\right).$$

It follows that the inverse function $\psi^{-1}$ is given by

$$x = \psi^{-1}(t) = \frac{2}{\sigma}\mathrm{erfc}^{-1}(2-2t) = \frac{2}{\sigma}\mathrm{erf}^{-1}(2t-1). \qquad (69)$$

## AUTHOR CONTRIBUTIONS

H.Q.M. formulated the mathematical framework and wrote the MATLAB code for Algorithm 1. M.S.B. and L.B. designed and performed all of the experiments. V.M. coordinated the project. All authors contributed to the writing of the manuscript.

## REFERENCES

[1] A. Angelova and S. Zhu, "Efficient object detection and segmentation for fine-grained recognition," in *CVPR*, 2013.

[2] V. Arsigny, P. Fillard, X. Pennec, and N. Ayache, "Geometric means in a novel vector space structure on symmetric positive-definite matrices," *SIMAX*, vol. 29, no. 1, 2007.

[3] R. Bhatia, *Positive Definite Matrices*. Princeton University Press, 2007.

[4] B. J. Boom, J. He, S. Palazzo, P. X. Huang, C. Beyan, H.-M. Chou, F.-P. Lin, C. Spampinato, and R. B. Fisher, "A research tool for long-term and continuous analysis of fish assemblage in coral-reefs using underwater camera footage," *Ecological Informatics*, vol. 23, pp. 83–97, 2014.

[5] P. Brodatz, *Textures: a photographic album for artists and designers*. New York: Dover Publications, 1966.

[6] B. Caputo, E. Hayman, and P. Mallikarjuna, "Class-specific material categorisation," in *Computer Vision, 2005. ICCV 2005. Tenth IEEE International Conference on*, vol. 2, Oct 2005, pp. 1597–1604 Vol. 2.

[7] Y. Chai, V. Lempitsky, and A. Zisserman, "Symbiotic segmentation and part localization for fine-grained categorization," in *ICCV*. IEEE, 2013, pp. 321–328.

[8] C.-C. Chang and C.-J. Lin, "LIBSVM: A library for support vector machines," *ACM TIST*, vol. 2, pp. 27:1–27:27, 2011.

[9] Z. Chebbi and M. Moakher, "Means of Hermitian positive-definite matrices based on the Log-Determinant $\alpha$-divergence function," *Linear Algebra and its Applications*, vol. 436, no. 7, pp. 1872–1889, 2012.

[10] G. Chen, J. Yang, H. Jin, E. Shechtman, J. Brandt, and T. Han, "Selective pooling vector for fine-grained recognition," in *WACV*, Jan 2015, pp. 860–867.

[11] A. Cherian, S. Sra, A. Banerjee, and N. Papanikolopoulos, "Jensen-Bregman LogDet divergence with application to efficient similarity search for covariance matrices," *PAMI*, vol. 35, no. 9, pp. 2161–2174, 2013.

[12] K. J. Dana, B. van Ginneken, S. K. Nayar, and J. J. Koenderink, "Reflectance and texture of real-world surfaces," *ACM Trans. Graph.*, vol. 18, no. 1, pp. 1–34, Jan. 1999.

[13] J. Dick, F. Kuo, and I. Sloan, "High-dimensional integration: the quasi-Monte Carlo way," *Acta Numerica*, vol. 22, pp. 133–288, 2013.

[14] J. Donahue, Y. Jia, O. Vinyals, J. Hoffman, N. Zhang, E. Tzeng, and T. Darrell, "Decaf: A deep convolutional activation feature for generic visual recognition," in *ICML*, 2014.

[15] D. Eigen, C. Puhrsch, and R. Fergus, "Depth map prediction from a single image using a multi-scale deep network," in *NIPS*, 2014, pp. 2366–2374.

[16] A. Ess, B. Leibe, and L. V. Gool, "Depth and appearance for mobile scene analysis," in *ICCV*, October 2007.

[17] M. Faraki, M. Harandi, and F. Porikli, "Approximate infinite-dimensional region covariance descriptors for image classification," in *ICASSP*, 2015.

[18] A. Feragen, F. Lauze, and S. Hauberg, "Geodesic exponential kernels: When curvature and linearity conflict," in *CVPR*, 2015.

[19] R. Girshick, J. Donahue, T. Darrell, and J. Malik, "Rich feature hierarchies for accurate object detection and semantic segmentation," in *CVPR*, 2014.

[20] M. Harandi, M. Salzmann, and F. Porikli, "Bregman divergences for infinite dimensional covariance matrices," in *CVPR*, 2014.

[21] B. Hariharan, P. Arbeláez, R. Girshick, and J. Malik, "Hypercolumns for object segmentation and fine-grained localization," in *CVPR*, 2015.

[22] S. Jayasumana, R. Hartley, M. Salzmann, H. Li, and M. Harandi, "Kernel methods on the Riemannian manifold of symmetric positive definite matrices," in *CVPR*, 2013.

[23] ——, "Kernel methods on Riemannian manifolds with Gaussian RBF kernels," *IEEE Transactions on Pattern Analysis and Machine intelligence*, vol. 37, no. 12, pp. 2464–2477, 2015.

[24] Y. Jia, E. Shelhamer, J. Donahue, S. Karayev, J. Long, R. Girshick, S. Guadarrama, and T. Darrell, "Caffe: Convolutional architecture for fast feature embedding," *arXiv*, 2014.

[25] A. Khosla, N. Jayadevaprakash, B. Yao, and L. Fei-Fei, "Novel dataset for fine-grained image categorization," in *CVPRW*, Colorado Springs, CO, June 2011.

[26] A. Krizhevsky, I. Sutskever, and G. E. Hinton, "Imagenet classification with deep convolutional neural networks," in *NIPS*, 2012, pp. 1097–1105.

[27] G. Kylberg, M. Uppstroem, K.-O. Hedlund, G. Borgefors, and I.-M. Sintorn, "Segmentation of virus particle candidates in transmission electron microscopy images," *Journal of Microscopy*, pp. no–no, 2011. [Online]. Available: http://dx.doi.org/10.1111/j.1365-2818.2011.03556.x

[28] G. Larotonda, "Nonpositive curvature: A geometrical approach to Hilbert Schmidt operators," *Differential Geometry and its Applications*, vol. 25, pp. 679–700, 2007.

[29] B. Leibe and B. Schiele, "Analyzing appearance and contour based methods for object categorization," in *CVPR*, vol. 2, June 2003, pp. II–409–15 vol.2.

[30] P. Li, Q. Wang, W. Zuo, and L. Zhang, "Log-Euclidean kernels for sparse representation and dictionary learning," in *ICCV*, 2013.

[31] Z. Liao, J. Rock, Y. Wang, and D. Forsyth, "Non-parametric filtering for geometric detail extraction and material representation," in *CVPR*, 2013, p. 963970.

[32] D. G. Lowe, "Distinctive image features from scale-invariant keypoints," *IJCV*, vol. 60, no. 2, pp. 91–110, 2004.

[33] H. Q. Minh, "Affine-invariant Riemannian distance between infinite-dimensional covariance operators," in *Geometric Science of Information*, 2015.

[34] H. Q. Minh, M. San Biagio, L. Bazzani, and V. Murino, "Approximate Log-Hilbert-Schmidt distances between covariance operators for image classification," in *The IEEE Conference on Computer Vision and Pattern Recognition (CVPR)*, June 2016.

[35] H. Minh, "Infinite-dimensional Log-Determinant divergences between positive definite trace class operators," *Linear Algebra and its Applications*, 2016, in press.

[36] H. Minh, M. San Biagio, and V. Murino, "Log-Hilbert-Schmidt metric between positive definite operators on Hilbert spaces," in *NIPS*, 2014.

[37] N. Murray and F. Perronnin, "Generalized max pooling," in *CVPR*, 2014.

[38] H. Niederreiter, *Random number generation and quasi-Monte Carlo methods*. SIAM, 1992.

[39] M.-E. Nilsback and A. Zisserman, "Automated flower classification over a large number of classes," in *ICVGIP*, Dec 2008.

[40] O. Parkhi, A. Vedaldi, A. Zisserman, and C. Jawahar, "Cats and dogs," in *CVPR*, 2012.

[41] X. Pennec, P. Fillard, and N. Ayache, "A Riemannian framework for tensor computing," *IJCV*, vol. 66, no. 1, pp. 41–66, 2006.

[42] F. Porikli, O. Tuzel, and P. Meer, "Covariance tracking using model update based on lie algebra," in *CVPR*, vol. 1. IEEE, 2006, pp. 728–735.

[43] A. Rahimi and B. Recht, "Random features for large-scale kernel machines," in *NIPS*, 2007.

[44] M. Reed and B. Simon, *Methods of Modern Mathematical Physics: Fourier analysis, self-adjointness*. Academic Press, 1975.

[45] A. Sharif Razavian, H. Azizpour, J. Sullivan, and S. Carlsson, "Cnn features off-the-shelf: An astounding baseline for recognition," in *CVPRW*, June 2014.

[46] S. Si, C.-J. Hsieh, and I. Dhillon, "Memory efficient kernel approximation," in *ICML*, 2014.

[47] K. Simonyan and A. Zisserman, "Very deep convolutional networks for large-scale image recognition," *ICLR*, 2015.

[48] S. Sra, "A new metric on the manifold of kernel matrices with application to matrix geometric means," in *Advances in Neural Information Processing Systems*, 2012, pp. 144–152.

[49] H. Tan, Z. Ma, S. Zhang, Z. Zhan, B. Zhang, and C. Zhang, "Grassmann manifold for nearest points image set classification," *Pattern Recognition Letters*, 2015.

[50] D. Tosato, M. Spera, M. Cristani, and V. Murino, "Characterizing humans on Riemannian manifolds," *PAMI*, vol. 35, no. 8, pp. 1972–1984, Aug 2013.

[51] O. Tuzel, F. Porikli, and P. Meer, "Pedestrian detection via classification on Riemannian manifolds," *PAMI*, vol. 30, no. 10, pp. 1713–1727, 2008.

[52] C. Wah, S. Branson, P. Welinder, P. Perona, and S. Belongie, "The Caltech-UCSD Birds-200-2011 Dataset," California Institute of Technology, Tech. Rep., 2011.

[53] R. Wang, H. Guo, L. Davis, and Q. Dai, "Covariance discriminative learning: A natural and efficient approach to image set classification," in *CVPR*, June 2012, pp. 2496–2503.

[54] A. Wendel and A. Pinz, "Scene categorization from tiny images," in *31st Annual Workshop of the Austrian Association for Pattern Recognition*, 2007, pp. 49–56.

[55] J. Yang, V. Sindhwani, H. Avron, and M. Mahoney, "Quasi-Monte Carlo feature maps for shift-invariant kernels," in *ICML*, 2014, pp. 485–493.

[56] N. Zhang, J. Donahue, R. Girshick, and T. Darrell, "Part-based r-cnns for fine-grained category detection," in *ECCV*, 2014, pp. 834–849.

[57] S. K. Zhou and R. Chellappa, "From sample similarity to ensemble similarity: Probabilistic distance measures in reproducing kernel Hilbert space," *PAMI*, vol. 28, no. 6, pp. 917–929, 2006.

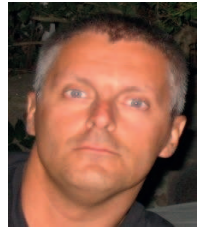

**Vittorio Murino** is full professor and head of the Pattern Analysis and Computer Vision (PAVIS) department at the Istituto Italiano di Tecnologia (IIT), Genoa, Italy. He received the Ph.D. in Electronic Engineering and Computer Science in 1993 at the University of Genoa, Italy. Then, he was first at the University of Udine and, since 1998, at the University of Verona, where he was chairman of the Department of Computer Science from 2001 to 2007. His research interests are in computer vision and machine learning, in particular, probabilistic techniques for image and video processing, with applications on video surveillance, biomedical image analysis and bio-informatics. He is also member of the editorial board of Pattern Recognition, Pattern Analysis and Applications, and Machine Vision & Applications journals, as well as of the IEEE Transactions on Systems Man, and Cybernetics. Finally, he is senior member of the IEEE and Fellow of the IAPR.

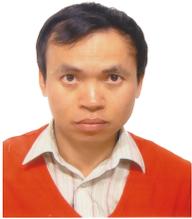

**Hà Quang Minh** received the Ph.D. degree in mathematics from Brown University, Providence, RI, USA, in May 2006. He is currently a Researcher in the Department of Pattern Analysis and Computer Vision (PAVIS) with the Istituto Italiano di Tecnologia (IIT), Genova, Italy. His current research interests include applied and computational functional analysis, machine learning, with applications in data analysis, computer vision, and image and signal processing.

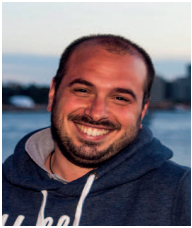

**Marco San Biagio** received the M.Sc. degree cum laude in Informatics Engineering from the University of Palermo, Italy, in 2010, and the Ph.D. in computer engineering from University of Genoa and Istituto Italiano di Tecnologia (IIT), Italy, in 2014, under the supervision of Prof. Vittorio Murino and Prof. Marco Cristani working on "Data Fusion in Video Surveillance". Before his current position, he was a postdoctoral fellow at the Pattern Analysis and Computer Vision department (PAVIS) in IIT, Genoa, Italy. His main research interests include machine learning, statistical pattern recognition and data fusion techniques for object detection and classification.

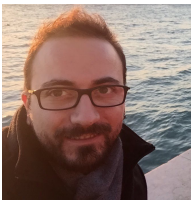

**Loris Bazzani** is currently a Computer Vision Scientist at Amazon in Berlin, Germany. He obtained the Ph.D. in Computer Science from the University of Verona (Italy) in 2012 supervised by Prof. Vittorio Murino. During his Ph.D., he spent 6 months at the University of British Columbia supervised by Prof. Nando de Freitas. Before the current position, he was a postdoctoral fellow at Dartmouth College working with Prof. Lorenzo Torresani and a postdoctoral fellow at the Italian Institute of Technology working with Prof. Vittorio Murino. His research is focused on person re-identification, social group analysis, object localization and detection and attentional models for computer vision.